\pdfoutput=1
\PassOptionsToPackage{table,x11names}{xcolor}

\documentclass[11pt,natbib]{article}

\usepackage[preprint]{coling}

\definecolor{lightblue}{RGB}{235,245,255}

\usepackage{times}
\usepackage{latexsym}

\usepackage[T1]{fontenc}

\usepackage[utf8]{inputenc}

\usepackage{microtype}

\usepackage{inconsolata}

\usepackage{graphicx}
\usepackage{algorithm}
\usepackage{algorithmic}
\usepackage{booktabs}
\usepackage{multirow}
\usepackage{amsmath}
\usepackage{newfloat}
\usepackage{listings}
\usepackage{makecell}
\usepackage{colortbl}
\usepackage[most]{tcolorbox}
\usepackage{tabularx}
\usepackage{placeins}
\usepackage{enumitem}

\title{Pedagogy-R1: Pedagogically-Aligned Reasoning Model with Balanced Educational Benchmark}

\author{
  \textbf{Unggi Lee\textsuperscript{1}$\dagger$},
  \textbf{Jaeyong Lee\textsuperscript{2}$\dagger$},
  \textbf{Jiyeong Bae\textsuperscript{1}},
  \textbf{Yeil Jeong\textsuperscript{2}},
  \textbf{Junbo Koh\textsuperscript{2}} \\
  \textbf{Gyeonggeon Lee\textsuperscript{3}},
  \textbf{Gunho Lee\textsuperscript{1}},
  \textbf{Taekyung Ahn\textsuperscript{1}},
  \textbf{Hyeoncheol Kim\textsuperscript{4}}
\\
  \textsuperscript{1}Enuma, Inc.,
  \textsuperscript{2}Seoul National University,
  \textsuperscript{3}Nanyang Technological University,
  \textsuperscript{4}Korea University
\\
  \small{
    \textbf{Corresponding Authors ($\dagger$):}
    \href{mailto:codingchild@korea.ac.kr}{codingchild@korea.ac.kr},
    \href{mailto:study@snu.ac.kr}{study@snu.ac.kr}
  }
}

\begin{document}
\maketitle

\begin{abstract}
Recent advances in large reasoning models (LRMs) show strong performance in structured domains like math and programming, but they lack pedagogical coherence and real‐world teaching behaviors. To bridge this gap, we introduce Pedagogy-R1, a framework that tailors LRMs for classroom use via three innovations: (1) a distillation‐based pipeline that filters and refines model outputs for instruction tuning, (2) the Well-balanced Educational Benchmark (WBEB), which measures performance across subject knowledge, pedagogy, tracing, essay scoring, and teacher decision–making, and (3) Chain-of-Pedagogy (CoP) prompts to generate and elicit teacher-style reasoning. Our mixed‐method evaluation combines quantitative metrics and qualitative analysis, offering the first systematic assessment of LRMs’ pedagogical strengths and limitations.
\end{abstract}

\section{Introduction}

In recent years, large language models (LLMs) have seen widespread adoption across various educational tasks, including adaptive tutoring, formative assessment and content generation \cite{kasneci2023chatgpt, yan2024practical, wang2024large, jeon2023large}. Despite this progress, there are still important limitations that hinder their effective use in real-world educational settings \cite{baidoo2023education, whalen2023chatgpt, rasul2023role}.

\begin{figure}[ht!]
    \centering
    \includegraphics[width=1\linewidth]{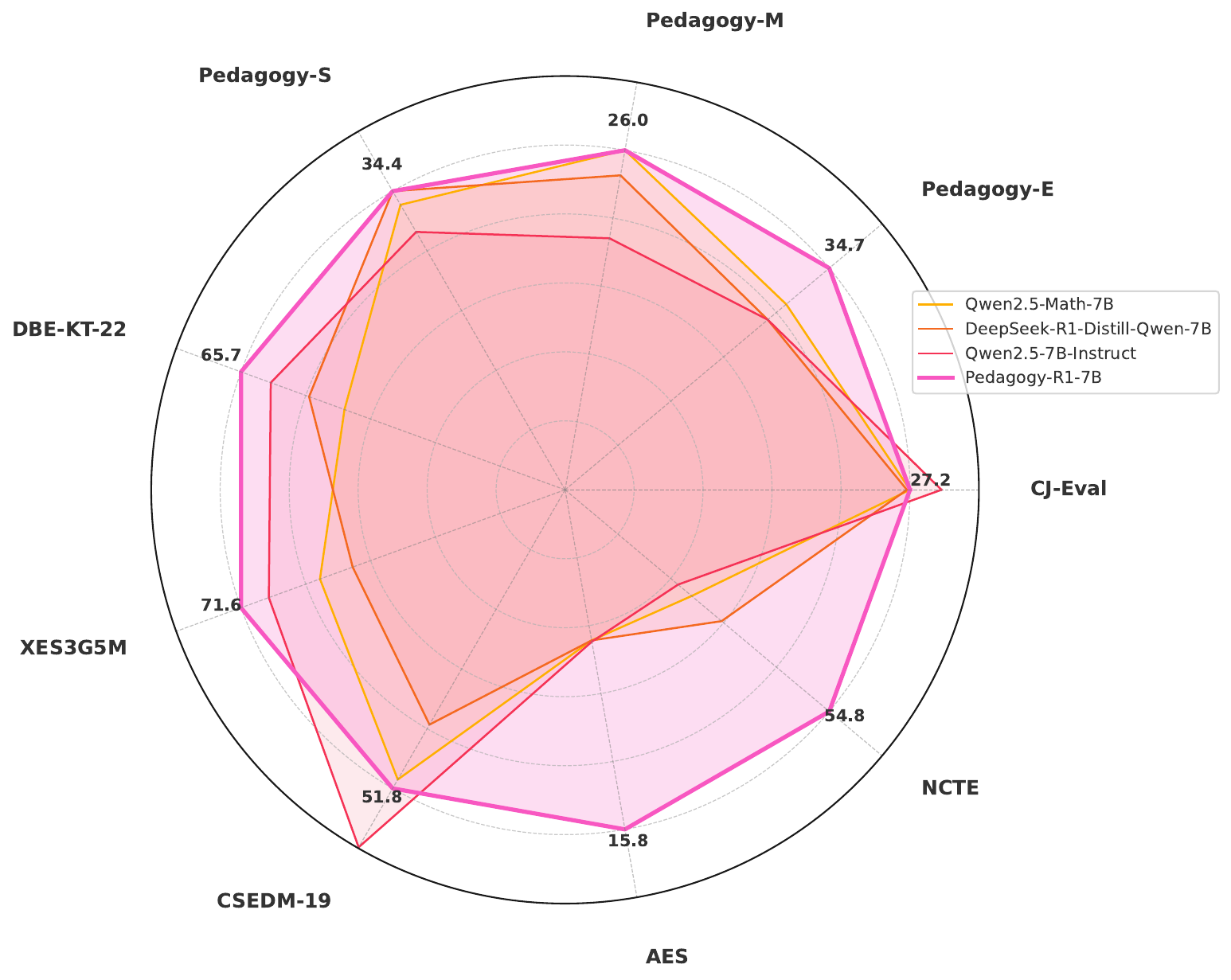}
    \caption{Radar chart illustrating the overall performance of our pedagogical reasoning model and baselines across all educational domains in our benchmark. The chart demonstrates the balanced and robust capabilities of our approach.}
    \label{fig:radar_chart_v1}
\end{figure}

First, LLMs in education mainly focus on answer generation and content delivery, often producing outputs that lack proper pedagogical alignment with real-world educational practices \cite{sonkar2024pedagogicalalignment, puech2025pedagogicalsteering}. In contrast, human teachers do not rely solely on content knowledge, but design instruction and guide students based on educational theories and informed pedagogical judgment \cite{schon2017reflective, borko2013teacher, eggleston2018teacher, siuty2018unraveling, doyle1977practicality}.

Second, existing benchmarks for LLMs in education focus mainly on content knowledge and overlook essential instructional skills required for real educational impact such as providing feedback, scaffolding student understanding, facilitating formative assessment, and managing classroom discourse \cite{zhang2024cjeval, jurenka2024learnlm}. Furthermore, benchmarks should reflect the diverse ways LLMs are deployed in educational settings—not only as adaptive tutors engaged in real-time dialogue with students, but also as background agents in automated grading systems, feedback generators, or components integrated within larger educational platforms \cite{fagbohun2024beyond, gao2024automatic, mizumoto2023exploring, fu2024sinkt, jung2024clst, lee2024language}. 

\begin{figure*}
    \centering
    \includegraphics[width=1\linewidth]{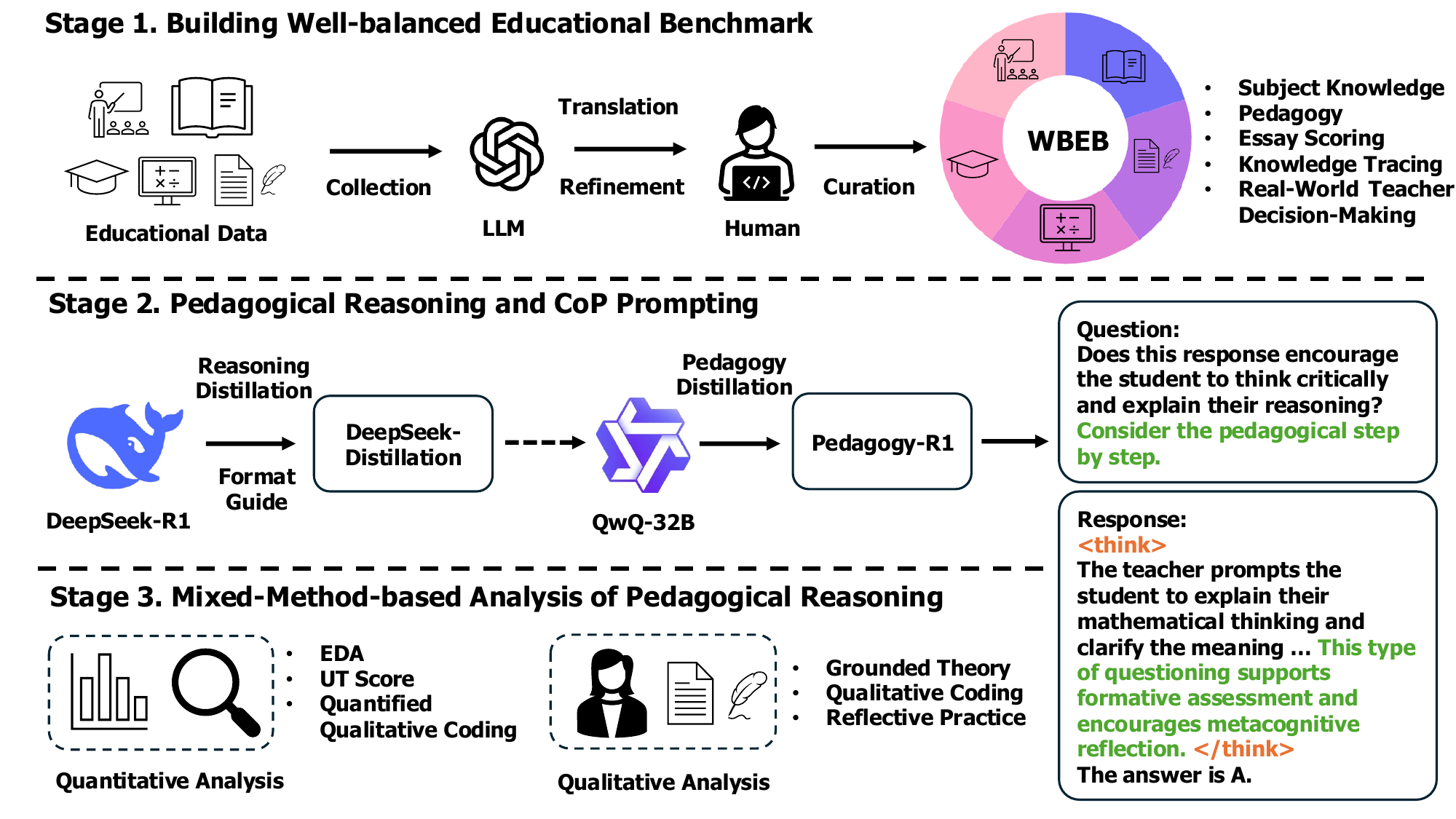}
    \caption{Overview of the development pipeline for the Well-balanced Educational Benchmark (WBEB) and Pedagogy-R1. The figure illustrates three key stages: (1) construction of a comprehensive educational benchmark through data collection, LLM translation, and human curation; (2) pedagogical reasoning via distillation and Chain-of-Pedagogy (CoP) prompting to train teacher-like models; and (3) mixed-method analysis combining quantitative and qualitative evaluation of pedagogical reasoning.}
    \label{fig:main_fig}
\end{figure*}

Third, existing LLMs do not reveal how they arrive at their answers, making it difficult to trace or evaluate their reasoning—a critical limitation in educational contexts where understanding the basis of responses is essential \cite{sonkar2024pedagogicalalignment, puech2025pedagogicalsteering, kasneci2023chatgpt, yan2024practical}.

Recently, large reasoning models (LRMs) have achieved strong performance in domains that require structured multi-step reasoning, such as mathematics, logic, and programming \cite{jaech2024openai, guo2025deepseek}.
LRMs have also been applied to challenging domains such as law, medicine, and science \cite{li2025system, xu2025towards, chen2025towards, griot2025large, yu2025evaluating, huang2025m1}.
However, their application to education remains largely unexplored, and there is limited understanding of their strengths and limitations, both from a research perspective and in terms of real-world educational practice. Our work addresses this gap by offering a systematic framework for evaluating and interpreting LRMs in educational research context.

To address these gaps, we propose a Pedagogical Reasoning, which trains LRMs to mimic the reasoning processes of human teachers and filters for pedagogically aligned outputs. This approach produces the Pedagogy-R1 model family, optimized for educational tasks.

For systematic evaluation, we introduce the Well-balanced Educational Benchmark (WBEB), covering five dimensions: subject knowledge (SK), pedagogical knowledge (PK), knowledge tracing (KT), automated essay scoring (AES), and classroom decision-making (DM).
We also develop the Chain-of-Pedagogy (CoP) prompting strategy to induce teacher-like reasoning and generate high-quality distillation data.

Finally, our analysis directly addresses the interpretability limitations of LLMs. Quantitatively, we examine patterns such as underthinking and apply domain-specific metrics. Qualitatively, we use grounded theory-based coding \cite{glaser2017discovery, glaser1998grounded, corbin2017grounded}, a widely used method in educational qualitative research \cite{nelson2020computational}, to analyze and compare model reasoning with teacher reasoning, drawing on Schön’s tripartite conception of professional reflection \cite{schon2017reflective} as our theoretical lens.

In summary, this study introduces a pedagogical reasoning framework and prompting method, and a comprehensive benchmark dataset, systematically analyzing both the potential and the limitations of LRMs in educational contexts.

\subsection{Contributions}

\begin{itemize}[leftmargin=*, itemsep=2pt, topsep=2pt, parsep=0pt]
    \item \textbf{Pedagogical Reasoning Framework}: We propose a distillation-based training pipeline that aligns LLMs with teacher-like reasoning through pedagogically filtered instruction tuning.
    \item \textbf{Chain-of-Pedagogy Prompting}: We develop CoP, a prompting strategy that elicits pedagogical reasoning both during teacher-model data generation and inference, enhancing educational alignment in both training and zero/few-shot use cases.
    \item \textbf{Well-balanced Educational Benchmark (WBEB)}: We introduce a comprehensive benchmark spanning five educational dimensions, incorporating both public and newly curated datasets to reflect real-world instructional needs.
    \item \textbf{Fine-grained Reasoning Analysis}: We provide in-depth reasoning evaluation through quantitative and qualitative analysis.
\end{itemize}

\section{Method}

\subsection{Pedagogical Reasoning}

To develop a pedagogical reasoning model, we adopted a distillation-based approach inspired by \texttt{DeepSeek-R1} \cite{guo2025deepseek}. The overall process involves generating high-quality instruction data using a teacher model and instruction-tuning student models to improve pedagogical reasoning.

\subsubsection{Model Selection for Data Generation}

We utilized the \texttt{QwQ-32B} \cite{qwq32b, qwen2.5} as the teacher model for data generation. This choice was motivated by its compatibility with the Qwen model family, ensuring architectural alignment with the distilled student models.

\subsubsection{Training Data Construction}

For training data generation, we used \(80\%\) of the WBEB benchmark dataset introduced in the next section. The remaining \(20\%\) of the dataset was reserved for evaluation purposes to ensure proper validation and testing of pedagogical reasoning performance.

Let \( D_{\text{train}} \subset D \) be the training portion of the dataset, such that:
\[
|D_{\text{train}}| = 0.8 \cdot |D|
\]

Each input \( x_i \in D_{\text{train}} \) was fed into the teacher model to obtain a generated response \( y_i \).

\subsubsection{Response Filtering}

Following the methodology of \texttt{DeepSeek-R1} \cite{guo2025deepseek} and \texttt{STILL-2} \cite{Slow_Thinking_with_LLMs_2}, we applied a filtering step to retain only high-quality responses. Specifically, we kept only the instances where the generated response \( y_i \) was correct, as verified by automatic evaluation or rule-based heuristics.

Let:
\[
D_{\text{filtered}} = \{ (x_i, y_i) \in D_{\text{train}} \mid \text{Correct}(y_i) = \text{True} \}
\]

\subsubsection{Instruction Tuning of Student Models}

We performed instruction tuning on two smaller models, \texttt{DeepSeek-R1-Distill-Qwen} \texttt{-1.5B} and \textit{DeepSeek-R1-Distill-Qwen-7B}, using the filtered dataset $D_{\text{filtered}}$. Both models are distilled variants of \texttt{DeepSeek-R1} \cite{guo2025deepseek}.

\subsubsection{Final Models}

The resulting models, \textit{Pedagogy-R1-1.5B} and \textit{Pedagogy-R1-7B}, were specifically optimized for tasks requiring pedagogical reasoning. These models aim to provide explanations, feedback, and instructional guidance more aligned with educational contexts.

\subsection{Chain-of-Pedagogy}

Inspired by Chain-of-Thought (CoT) prompting, we propose a pedagogically grounded variant called \textit{Chain-of-Pedagogy} (CoP) (Figure~\ref{fig:prompt_evolution}). While commonly used prompts such as \texttt{Please reason step by step, and then pick the best option} are effective for general reasoning, they fall short of eliciting pedagogically meaningful responses—those that reflect teaching intent, learner consideration, or instructional scaffolding.

To bridge this gap, we designed prompts that explicitly instruct the model to reason with pedagogical awareness. For example, we explored variants to foreground the educational context in the reasoning process.

\begin{figure}[h]
\centering
\begin{tcolorbox}[colback=gray!5!white, colframe=black!70, 
                  width=\columnwidth, boxrule=0.4pt, arc=2mm, 
                  left=1mm, right=1mm, top=1mm, bottom=1mm]
\texttt{Please reason step by step, and then pick the best option.} \\[0.2em]
\centerline{$\downarrow$} \\[0.2em]
\texttt{Consider the \textbf{\textcolor{blue}{pedagogical}} step by step, and then pick the best option.}
\end{tcolorbox}
\caption{Prompt refinement from general step-by-step reasoning to pedagogically guided reasoning.}
\label{fig:prompt_evolution}
\end{figure}

We integrate CoP in two core stages of our framework. First, we apply CoP prompting during inference time to guide the model’s step-by-step reasoning toward educationally aligned outputs. Second, we leverage CoP prompts during data generation by the teacher model—instructing it to produce pedagogically rich responses based on seed questions. These curated outputs are then used for distillation into smaller student models, helping them internalize structured pedagogical reasoning patterns. 

To identify the most effective prompt formulations, we employed a human-in-the-loop evaluation process. Using GPT-4o, we generated a diverse set of CoP candidates and manually reviewed dozens of outputs per prompt. Rather than relying solely on quantitative metrics, we prioritized interpretability—how clearly the model expressed pedagogical strategies such as scaffolding, formative feedback, or instructional decision-making.

\begin{table*}[t]
\centering
\small
\caption{Summary of WBEB datasets by category, key stats, and availability for model evaluation.}
\resizebox{\textwidth}{!}{%
\begin{tabular}{clrrcccc}
\toprule
\textbf{Categories} & \textbf{Dataset Name} & \textbf{\# Samples} & \textbf{Avg. len} & \textbf{Task Type} & \textbf{Lang} & \textbf{Availability} & \textbf{Usage} \\
\midrule
\textbf{\makecell[c]{\textbf{Subject}\\\textbf{Knowledge}\\\textbf{(SK)}}}
    & cj-eval-T \cite{zhang2024cjeval} & 805 & 65,985 & MCQA & Ch -> En & Public & Research Only \\
\midrule
\multirow{3}{*}{\makecell[c]{\textbf{Pedagogical}\\\textbf{Knowledge}\\\textbf{(PK)}}}
    & kice-pedagogy-elementary & 124 & 15,959 & MCQA & Ko -> En & Private & Internal Use Only \\
    & kice-pedagogy-middle & 104 & 16,512 & MCQA & Ko -> En & Private & Internal Use Only \\
    & kice-pedagogy-middle-special & 64 & 6,036 & MCQA & Ko -> En & Private & Internal Use Only \\
\midrule
\multirow{6}{*}{\makecell[c]{\textbf{Knowledge}\\\textbf{Tracing}\\\textbf{(KT)}}} 
    & dbe-kt-22 (zeroshot) & 3,146 & 3,637,585 & BC & En & Public & Research Only \\
    & dbe-kt-22 (fiveshot) & 3,146 & 18,105,770 & BC & En & Public & Research Only \\
    & xes3g5fm-simple (zeroshot) & 5,000 & 4,980,121 & BC & Ch -> En & Public & Research/Commercial \\
    & xes3g5fm-simple (fiveshot) & 5,000 & 23,502,103 & BC & Ch -> En & Public & Research/Commercial \\
    & csedm19-spring (zeroshot) & 3,992 & 3,913,452 & BC & Code -> En & Public & Research Only \\
    & csedm19-spring (fiveshot) & 3,992 & 18,458,602 & BC & Code -> En & Public & Research Only \\
\midrule
\textbf{\makecell[c]{Automated\\Essay Scoring\\(AES)}} 
    & learning-agency-lab-aes & 3,462 & 1,588,192 & OC & En & Public & Research/Commercial \\
\midrule
\multirow{5}{*}{\makecell[c]{\textbf{Real-World}\\\textbf{Teacher}\\\textbf{Decision-Making}\\\textbf{(DM)}}} 
    & ncte-eval-mcq-aa & 471 & 100,327 & MCQA & En & Private & Research Only \\
    & ncte-eval-bn-fq & 471 & 47,560 & BC & En & Private & Research Only \\
    & ncte-eval-bn-hu & 471 & 49,564 & BC & En & Private & Research Only \\
    & ncte-eval-bn-sr & 401 & 19,813 & BC & En & Private & Research Only \\
    & ncte-eval-mcq-sr & 85 & 9,487 & MCQA & En & Private & Research Only \\
\bottomrule
\end{tabular}
}
\label{tab:edu_benchmark}
\end{table*}

\section{Well-balanced Educational Benchmark}

Existing benchmarks for evaluating LLMs in education have primarily focused on assessing subject-specific content knowledge. However, in real-world educational settings, LLMs are increasingly expected to perform a broader range of pedagogical functions—providing student feedback, scaffolding learning, grading open-ended responses, and supporting in-situ instructional decisions. Narrow evaluation criteria risk overlooking these critical capabilities.

To address this gap, we designed a benchmark that systematically evaluates LLMs across five pedagogically grounded categories: \texttt{subject knowledge (SK)}, \texttt{pedagogical knowledge (PK)}, \texttt{knowledge tracing (KT)}, \texttt{automated essay scoring (AES)}, and \texttt{real-world teacher decision-making (DM)}. Each category (Table \ref{tab:edu_benchmark}) is represented by datasets that reflect authentic educational needs, combining well-established public resources with newly curated or transformed data to ensure coverage, diversity, and practical relevance.

\subsection{Subject Knowledge}

To evaluate how well LLMs support subject-specific reasoning, we utilize the \texttt{cj-eval-T} dataset, a translated subset of CJ-Eval \cite{zhang2024cjeval} derived from authentic Chinese junior high school exam questions. The original questions were written in Chinese and translated into English using GPT-4.1-mini to ensure consistent evaluation across models.

To maintain compatibility with the overall benchmark structure, we selected only the multiple choice question answering (MCQA) items from the dataset, excluding other formats such as fill-in-the-blank or analytical questions. Each item includes a question stem, four to five options, and one correct answer. The dataset spans ten academic subjects and includes metadata such as knowledge concepts and empirically derived difficulty levels, categorized into five tiers based on student error rates.

The \texttt{cj-eval-T} dataset is publicly available for research purposes only. It provides a standardized and linguistically consistent resource to assess the subject knowledge capabilities of language models across diverse disciplines.

\subsection{Pedagogical Knowledge}

To assess LLMs' understanding of core pedagogical concepts and practices, we construct the \texttt{kice-pedagogy} series using multiple-choice items from the official teacher employment certification exams administered by the Korea Institute for Curriculum and Evaluation (KICE). These national-level exams are available via the KICE archive\footnote{\url{https://www.kice.re.kr/boardCnts/list.do?boardID=1500212&m=030306&s=kice}}, but are subject to strict copyright restrictions prohibiting reproduction, redistribution, or derivative use.

We selected elementary education items from the years 2001–2012, secondary education items from 2003–2013, and special education items from 2003–2008. Only MCQA were included to ensure format consistency with other benchmark components. All items were originally written in Korean.

To build the dataset, we collected official exam PDFs from the KICE website and applied OCR to extract the text. OCR errors, especially in older documents, were manually reviewed and corrected to ensure fidelity. The cleaned Korean texts were then translated into English using GPT-4.1-mini, with additional human verification for accuracy and conceptual alignment.

Due to copyright limitations, this dataset is used strictly for internal experimentation. We do not release the data or any models trained on it. Publicly released models are trained without this dataset to ensure full compliance.

\begin{table*}[t]
\centering
\scriptsize
\caption{Performance comparison of all models across WBEB domains. Pedagogical Reasoning models—especially with CoP Prompting—consistently achieve the best results across all domains except for SK. CoP distillation sometimes improves performance but not uniformly across tasks.}
\resizebox{\textwidth}{!}{%
\begin{tabular}{l|cccccc}
\toprule
\textbf{Model Name} & \textbf{SK (ACC)} & \textbf{PK (ACC)} & \textbf{KT (AUC)} & \textbf{KT (ACC)} & \textbf{AES (ACC)} & \textbf{DM (ACC)} \\
\midrule
\rowcolor{lightgray} \multicolumn{7}{c}{\textbf{Base Models}} \\
\midrule
Qwen2.5-Math-1.5B & 26.83 & 27.84 & 54.77 & 59.40 & 33.02 & 27.31 \\
Qwen2.5-Math-7B   & 27.20 & 29.27 & 50.00 & 54.43 & 7.02 & 26.24 \\
Llama-3.1-8B-Instruct & 31.80 & 28.80 & 50.29 & 59.40 & 7.02 & 52.41 \\
Qwen2.5-14B & 36.02 & 29.91 & 58.81 & 62.14 & 7.02 & 42.40 \\
Qwen2.5-32B & 35.65 & 32.31 & 64.73 &  72.20 & 7.02 & 45.80 \\
Llama-3.3-70B-Instruct & 37.14 & 31.69 & 56.49 & 59.40 & 7.02 & 54.67 \\
\midrule
\rowcolor{lightgray} \multicolumn{7}{c}{\textbf{Base Models (Reasoning)}} \\
\midrule
DeepSeek-R1-Distill-Qwen-1.5B & 23.98 & 28.58 & 50.89 & 59.40 & 7.08 & 24.68 \\
DeepSeek-R1-Distill-Qwen-7B & 26.96 & 28.34 & 49.06 & 58.39 & 7.02 & 32.45 \\
DeepSeek-R1-Distill-Llama-8B & 28.20 & 25.15 & 51.54 & 57.82 & 7.02 & 54.35 \\
DeepSeek-R1-Distill-Qwen-14B & 31.80 & 27.52 & 52.05 & 59.41 & 7.02 & 52.28 \\
DeepSeek-R1-Distill-Qwen-32B & 31.02 & 27.87 & 51.22 & 59.40 & 7.02 & 55.21 \\
DeepSeek-R1-Distill-Llama-70B & 31.93 & 28.82 & 56.88 & 59.40 & 7.02 & 56.06 \\
o4-mini & 45.96 & 50.09 & 60.15 & 76.58 & 20.05 & 65.52 \\
\midrule
\rowcolor{lightgray} \multicolumn{7}{c}
{\textbf{Instruction-Tuning on Our Data}} \\
\midrule
Qwen2.5-1.5B-Instruct & 25.34 & 24.26 & 49.57 & 40.60 & 7.02 & 23.51 \\
Qwen2.5-7B-Instruct & 29.69 & 25.18 & 59.81 & 40.60 & 7.08 & 23.42 \\
\midrule
\rowcolor{lightblue} \multicolumn{7}{c}{\textbf{Pedagogical Reasoning (Ours)}} \\
\midrule
Pedagogy-R1-1.5B (Ours) & 
23.60 \textcolor{red!60!black}{\scriptsize (-1.74)} & 
30.60 \textcolor{green!60!black}{\scriptsize (+6.34)} & 
52.57 \textcolor{green!60!black}{\scriptsize (+3.00)} & 
40.63 \textcolor{green!60!black}{\scriptsize (+0.03)} & 
7.02 \textcolor{green!60!black}{\scriptsize (+0.00)} & 
24.73 \textcolor{green!60!black}{\scriptsize (+1.22)} \\
Pedagogy-R1-7B (Ours) & 
27.20 \textcolor{red!60!black}{\scriptsize (-2.49)} & 
31.67 \textcolor{green!60!black}{\scriptsize (+6.49)} & 
54.99 \textcolor{red!60!black}{\scriptsize (-4.82)} & 
52.03 \textcolor{green!60!black}{\scriptsize (+11.43)} & 
15.83 \textcolor{green!60!black}{\scriptsize (+8.75)} & 
54.76 \textcolor{green!60!black}{\scriptsize (+31.34)} \\
\midrule
\rowcolor{lightblue} \multicolumn{7}{c}{\textbf{Pedagogical Reasoning + CoP Prompting (Ours)}} \\
\midrule
Pedagogy-R1-1.5B + CoP (Ours) & 
23.60 \textcolor{red!60!black}{\scriptsize (-1.74)} & 
30.88 \textcolor{green!60!black}{\scriptsize (+6.62)} & 
51.89 \textcolor{green!60!black}{\scriptsize (+2.23)} & 
40.83 \textcolor{green!60!black}{\scriptsize (+0.23)} & 
7.02 \textcolor{green!60!black}{\scriptsize (+0.00)} & 
24.96 \textcolor{green!60!black}{\scriptsize (+1.45)} \\
Pedagogy-R1-7B + CoP (Ours) & 
27.58 \textcolor{red!60!black}{\scriptsize (-2.21)} & 
31.13 \textcolor{green!60!black}{\scriptsize (+5.95)} & 
55.57 \textcolor{red!60!black}{\scriptsize (-4.24)} & 
46.35 \textcolor{green!60!black}{\scriptsize (+5.57)} & 
16.23 \textcolor{green!60!black}{\scriptsize (+9.15)} & 
54.80 \textcolor{green!60!black}{\scriptsize (+31.38)} \\
\midrule
\rowcolor{lightblue} \multicolumn{7}{c}{\textbf{Pedagogical Reasoning (CoP Distillation) (Ours)}} \\
\midrule
Pedagogy-R1-CoP-1B (Ours) & 
23.98 \textcolor{red!60!black}{\scriptsize (-1.36)} & 
30.67 \textcolor{green!60!black}{\scriptsize (+6.47)} & 
50.04 \textcolor{red!60!black}{\scriptsize (-0.47)} & 
38.68 \textcolor{red!60!black}{\scriptsize (-1.92)} & 
7.02 \textcolor{green!60!black}{\scriptsize (+0.00)} & 
28.45 \textcolor{green!60!black}{\scriptsize (+4.94)} \\

Pedagogy-R1-CoP-7B (Ours) & 
27.45 \textcolor{red!60!black}{\scriptsize (-2.24)} & 
31.13 \textcolor{green!60!black}{\scriptsize (+5.95)} & 
57.91 \textcolor{red!60!black}{\scriptsize (-1.9)} & 
73.71 \textcolor{green!60!black}{\scriptsize (+33.11)} & 
7.02 \textcolor{red!60!black}{\scriptsize (-0.06)} & 
36.46 \textcolor{green!60!black}{\scriptsize (+13.04)} \\

\midrule
\rowcolor{lightblue} \multicolumn{7}{c}{\textbf{Pedagogical Reasoning (CoP Distillation) + CoP Prompting (Ours)}} \\
\midrule
Pedagogy-R1-CoP-1B + CoP (Ours) & 
23.85 \textcolor{red!60!black}{\scriptsize (-1.49)} & 
29.77 \textcolor{green!60!black}{\scriptsize (+5.51)} & 
49.09 \textcolor{red!60!black}{\scriptsize (-0.48)} & 
39.99 \textcolor{red!60!black}{\scriptsize (-0.61)} & 
7.05 \textcolor{green!60!black}{\scriptsize (+0.03)} & 
25.89 \textcolor{green!60!black}{\scriptsize (+2.38)} \\
Pedagogy-R1-CoP-7B + CoP (Ours) & 
27.08 \textcolor{red!60!black}{\scriptsize (-2.61)} & 
29.70 \textcolor{green!60!black}{\scriptsize (+4.52)} & 
57.36 \textcolor{red!60!black}{\scriptsize (-2.45)} & 
63.30 \textcolor{green!60!black}{\scriptsize (+22.7)} & 
7.02 \textcolor{red!60!black}{\scriptsize (-0.06)} & 
25.91 \textcolor{green!60!black}{\scriptsize (+2.49)} \\
\bottomrule
\end{tabular}
}
\label{tab:performance}
\end{table*}

\subsection{Knowledge Tracing}

Knowledge Tracing (KT) is the task of modeling and predicting a student's evolving knowledge state over time based on their past interactions. Despite recent advances in LLMs, KT remains a challenging task. While models may achieve moderate performance in terms of accuracy (ACC), they consistently underperform on the area under the ROC curve (AUC)—the primary metric for evaluating KT systems—highlighting the gap between surface-level correctness prediction and deeper modeling of learning dynamics. For each dataset, we prepared both zero-shot and five-shot evaluation settings.

\subsubsection{dbe-kt22}
The \texttt{dbe-kt22} \cite{abdelrahman2022dbe} contains university-level relational database exercises from online student evaluations at the Australian National University. Each student's sequence is formed from item–response interactions, and models predict correctness in a zero-shot setting. The dataset is available for research use only.

\subsubsection{xes3g5m-simple}
The \texttt{xes3g5m} \cite{liu2024xes3g5m} is large-scale dataset covers third-grade math performance originally collected in Chinese and translated to English using GPT-4o-mini. To adapt for LLMs, we structured it as one sequence per student. For efficient evaluation, we provide \texttt{xes3g5m-simple}, a sampled subset of 5,000 representative sequences. The dataset is released under the MIT license, supporting both research and commercial use.

\subsubsection{csedm19-spring}
This dataset is tailored for programming education and was released by the CSEDM Workshop at LAK 2019 \cite{CSEDM2019}. Since the original questions and knowledge concepts were unavailable, we generated them from student-submitted Java code using GPT-4o-mini. Each interaction includes a reconstructed question, a generated knowledge concept, and correctness labels. The dataset supports evaluating KT performance in programming contexts and is available for research use only.

\subsection{Automated Essay Scoring}

This category focuses on evaluating open-ended student responses, a critical component of real-world formative assessment. The \texttt{aes} dataset originates from the 2024 Kaggle competition “Learning Agency Lab - Automated Essay Scoring 2.0,” hosted in collaboration with Vanderbilt University and The Learning Agency Lab \cite{crossley2024aes}. It includes large open-access dataset of student essays aligned with classroom writing tasks and national educational standards. The dataset is publicly available for both research and commercial use.

\subsection{Real-World Teacher Decision-Making}

This category evaluates how LLMs interpret authentic classroom discourse and instructional decision-making. The \texttt{ncte-eval} dataset \cite{demszky2023ncte}, drawn from the National Center for Teacher Effectiveness (NCTE) study, comprises 1,660 anonymized transcripts of 4th and 5th grade math lessons recorded between 2010 and 2013 in U.S. public schools serving historically marginalized communities.

Each transcript captures full-length classroom interactions and is enriched with metadata such as turn-level annotations (e.g., teacher uptake, focusing questions, student reasoning), expert observation scores (CLASS, MQI), teacher and student demographics, value-added scores, and survey responses. This combination allows for robust analysis of pedagogical reasoning and instructional quality.

Although the dataset is not publicly released, it is available for research use under data-sharing agreements. It offers a unique benchmark for assessing how well LLMs can model in-situ teaching practices and align with expert-rated measures of classroom effectiveness.

\section{Experiments}

\subsection{Experiment Setup}

To evaluate the effectiveness of our pedagogical reasoning framework, we compared four model groups, as summarized in Table~\ref{tab:performance}. First, we include standard base and their reasoning models (e.g., \texttt{\detokenize{DeepSeek-R1-Distill}}) which serve as strong general-purpose and reasoning-aware baselines. Second, we trained instruction-tuned models—\texttt{\detokenize{Qwen2.5-1.5B-Instruct}} and \texttt{\detokenize{Qwen2.5-7B-Instruct}} using only problem–answer pairs sampled from our training set. Finally, we evaluated our proposed \texttt{\detokenize{Pedagogy-R1}} models.

For the \texttt{\detokenize{Pedagogy-R1}}, we used \texttt{\detokenize{QwQ-32B}} as the teacher model to generate synthetic data based on 5,000 seed samples uniformly sampled across all benchmark categories. We retained only those instances where the generated answers were correct, following a filtering step aligned with prior distillation methods such as \texttt{\detokenize{DeepSeek-R1}} and \texttt{\detokenize{STILL-2}}. This resulted in a training dataset of 1,948 high-quality examples used to fine-tune two model variants: \texttt{\detokenize{Pedagogy-R1-1.5B}} and \texttt{\detokenize{Pedagogy-R1-7B}}.

Note that the instruction-tuned models \texttt{\detokenize{Qwen2.5-1.5B-Instruct}} and \texttt{\detokenize{Qwen2.5-7B-Instruct}} were also trained on the same 1,948 samples, but using only the problem and the correct answer without pedagogical reasoning. This distinction allows for isolating the impact of pedagogical reasoning in the training objective.

\subsection{Results}

Table~\ref{tab:performance} presents the performance of all models across six dimensions of the WBEB. Overall, the Pedagogy-R1 models show balanced and effective performance across educationally aligned tasks. Pedagogy-R1-7B achieves the highest score in DM (54.76\%), PK (31.67\%), and AES (15.83\%), with comparable KT performance (AUC: 54.99\%, ACC: 52.03\%). Pedagogy-R1-1.5B also performs competitively, especially in PK (30.60\%) and KT-AUC (52.57\%). Although these models do not always outperform in SK (27.20\% for 7B, 23.60\% for 1.5B), their advantage is more pronounced in dimensions requiring pedagogical reasoning and instructional decision-making. Detailed results for PK subdomains are presented in Table~\ref{tab:pk_subdomains}.

\begin{table}[h]
\centering
\scriptsize
\caption{Performance of models on PK subdomains: general elementary education (Elementary), general middle school education (Middle), and special education in middle school (Middle Special).}
\resizebox{\columnwidth}{!}{%
  \begin{tabular}{lccc}
    \toprule
    \textbf{Model} & \textbf{Elementary} & \textbf{Middle} & \textbf{Middle Special} \\
    \midrule
    Qwen2.5-1.5B-Instruct     & 0.2661 & 0.2115 & 0.2500 \\
    Pedagogy-R1-1.5B          & 0.2742 & \textbf{0.2692} & 0.3594 \\
    Pedagogy-R1-1.5B + CoP    & \textbf{0.2823} & \textbf{0.2692} & \textbf{0.3750} \\
    \bottomrule
  \end{tabular}%
}
\label{tab:pk_subdomains}
\end{table}

Compared to instruction-tuned models trained only on problem-answer pairs (Qwen2.5-1.5B-Instruct and Qwen2.5-7B-Instruct), the Pedagogy-R1 models consistently demonstrate improved alignment in pedagogical dimensions. Pedagogy-R1-1.5B outperforms Qwen2.5-1.5B-Instruct by +6.34\%p in PK (30.60\% vs. 24.26\%) and +3.00\%p in KT-AUC (52.57\% vs. 49.57\%). Pedagogy-R1-7B exceeds Qwen2.5-7B-Instruct by +6.49\%p in PK (31.67\% vs. 25.18\%), +11.43\%p in KT-ACC (52.03\% vs. 40.60\%), +8.75\%p in AES (15.83\% vs. 7.08\%), and +31.34\%p in DM (54.76\% vs. 23.42\%). These results show that pedagogically filtered instruction tuning leads to better performance in educational reasoning tasks than simple instruction tuning based on problem-answer formats.

Among the base models, performance trends partially correlate with model size. In the Qwen2.5 series, SK increases from 26.83\% (1.5B) to 35.65\% (32B), and KT-AUC increases from 54.77\% (1.5B) to 64.73\% (32B). Similarly, Llama-3.3-70B-Instruct achieves relatively strong performance in SK (37.14\%) and DM (54.67\%). In the DeepSeek-R1-Distill models, the Llama-70B variant records the highest DM (56.06\%) and KT-AUC (56.88\%) scores among its group. However, these gains are less consistent in PK and AES, where smaller pedagogically aligned models like Pedagogy-R1 outperform them. This suggests that while larger model size contributes to certain capability gains, pedagogical fine-tuning remains critical for educational reasoning and alignment tasks.

\section{Empirical Reasoning Analysis}

\subsection{Quantitative Analysis}

\begin{figure*}
    \centering
    \includegraphics[width=1\linewidth]{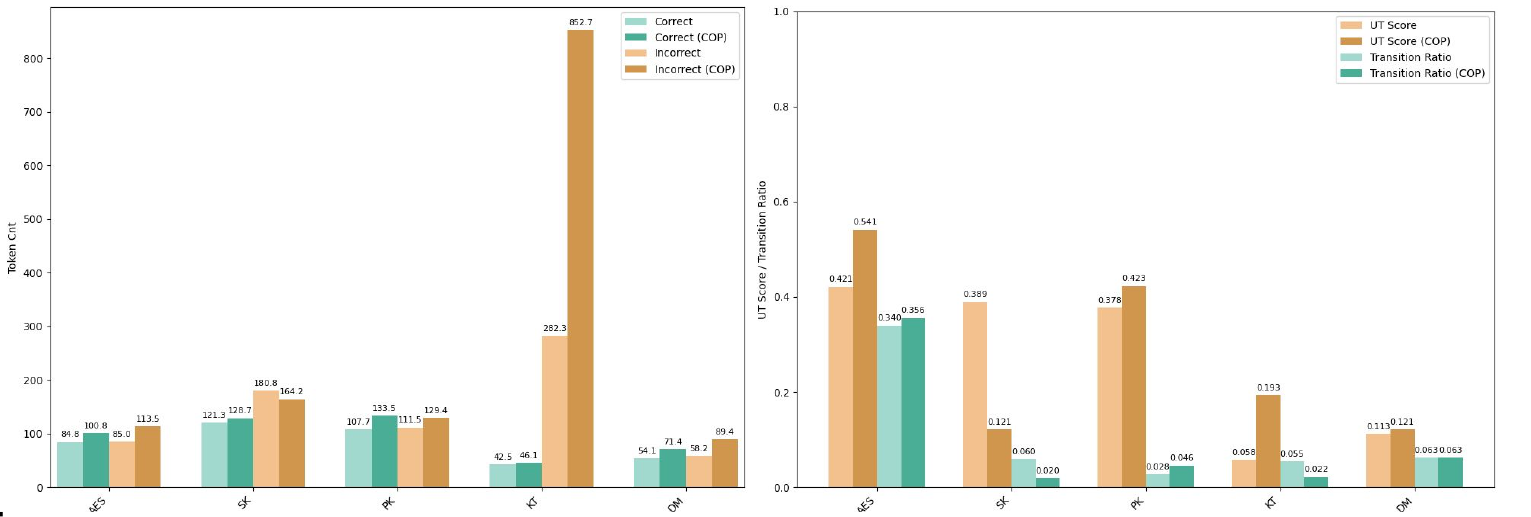}
    \captionsetup{skip=1pt}
    \caption{Quantitative analyses of domain-specific data. \textit{Left} shows reasoning token amounts, and \textit{Right} shows UT scores and transition word ratios.}
    \label{fig:analysis_1.1_fig}
\end{figure*}

We performed a comprehensive quantitative analysis of model reasoning patterns across multiple educational domains. The \textit{Left} panel of \ref{fig:analysis_1.1_fig} compares the amount of reasoning tokens for correct (\textit{yellow}) and incorrect (\textit{green}) answers across domains. Overall, most domains show higher reasoning token counts when the answer is incorrect. Additionally, COP prompting generally increases the token counts regardless of answer correctness.

When comparing domains, KT and SK exhibit the highest reasoning token counts, while DM shows the lowest. In particular, KT shows a significant increase in reasoning tokens for incorrect answers, possibly due to the task's complexity in next-token prediction tasks. CoP prompting further increases the reasoning token in KT for incorrect responses. Notably, Table \ref{tab:performance} shows that SK's performance drops, and it also consumes more tokens in incorrect cases. Importantly, prior studies using LLMs for KT \cite{jung2024clst, lee2024language, zhan2024knowledge, lee2024difficulty}, which rely on next token prediction, have typically reported AUC scores around 0.5 to 0.6 rather low performance level that highlights the fundamental challenge of this approach. Even when reasoning-augmented models were applied to next token prediction in KT of our research, they struggled to achieve meaningful improvements over standard methods.

The \textit{Right yellow} shows the underthikning (UT) Score \cite{wang2025thoughts}, which measures the proportion of unnecessary tokens generated after reaching the correct thought during incorrect reasoning. A higher score indicates less efficient inference. AES, SK, and PK exhibit high UT Scores, while KT and DM show lower values, indicating more efficient reasoning. Notably, KT, despite high reasoning token usage (\textit{Left}), maintains a lower UT Score, which will be further discussed qualitatively. In SK, CoP prompting reduces both reasoning tokens and UT Score, highlighting the effectiveness of pedagogical prompting. DM consistently shows both low reasoning tokens and UT Score, suggesting minimal token waste.

Additionally, following \cite{wang2025thoughts} and \cite{chen2025seal}, we observed that usage of contrast transition words correlates with higher UT Scores across domains (\textit{Right green}). However, in KT, COP prompting sometimes increases both reasoning tokens and UT Score but reduces contrast transitions, indicating a complex interplay between prompting and transition strategies.

\subsection{Qualitative Analysis}

To probe how the CoP prompt shaped the model’s reasoning, we adopted Schön’s  tripartite conception of professional reflection \cite{schon2017reflective} (reflection-in-action (IA), reflection-on-action (OA), and reflection-for-action / triggers (T)), as our provisional theoretical lens. The unit of analysis was the thought-token trace produced by the LLM while completing 42 item–response tasks.

Two researchers independently open-coded a stratified sample comprising approximately 33 \% of the LLM thought-token traces (14 of 42) using our custom a priori codebook, which was developed for this dataset to map reasoning tokens to reflection codes in real time. Through negotiated consensus the researchers reconciled differences and refined operational definitions, producing a stable codebook of ten codes (e.g., IA-CC criteria check, OA-SR summative reflection, T-C conflict cue). The codebook, developed to map LLM thought tokens to codes in real time, enables an explainable format of the model’s reasoning that is readily interpretable by end-users. The remaining 66 \% of traces were single-coded with periodic peer debriefing.

\begin{table}[htbp]
  \caption{Codes used in the evaluation process}
  \centering
  \small
  \begin{tabularx}{\columnwidth}{@{} l l X @{}}
    \toprule
    \textbf{Code} & \textbf{Code Meaning} & \textbf{Description} \\
    \midrule
    IA-SQ & IA-Self-Questioning     & Raising questions to oneself during the evaluation \\
    IA-CC & IA-Criteria-Check       & Checking against rubric criteria \\
    IA-IA & IA-Immediate-Adjustment & Modifying scores or feedback during the thinking process \\
    IA-P  & IA-Planning             & Planning the evaluation \\
    IA-C  & IA-Clarification        & Clarifying the meaning \\
    OA-SR & OA-Summative-Reflection & Reflecting on the entire evaluation process after its completion \\
    OA-ME & OA-Meta-Evaluation      & Evaluating the quality of the evaluation itself \\
    T-U   & T-Uncertainty           & Recognizing lack of information or ambiguity \\
    T-C   & T-Conflict              & Recognizing conflict between criteria \\
    \bottomrule
  \end{tabularx}
  \label{tab:evaluation_codes}
\end{table}

\subsubsection{Theme 1: Absence of Metacognitive Self-Critique}
Across all 42 traces produced before the CoP prompt was applied, no instances of IA-SQ (self-questioning) or OA-ME (meta-evaluation) codes were observed. After introducing the CoP prompt, these metacognitive codes began to appear, albeit sparsely, indicating that pedagogical prompting can elicit self-reflection that vanilla prompting fails to trigger. This pattern suggests that ordinary prompts are insufficient for inducing genuine self-critique, whereas CoP prompting partially alleviates the gap. To strengthen metacognition, future work might incorporate explicit self-audit instructions or link a first-pass evaluator model to a second-pass critic model.

\subsubsection{Theme 2: Recurrence of Filler or 'Useless' tokens}
Approximately 65 \% of all tokens were classified as noise: verbatim repetitions of the stem of the item or generic phatic remarks. These strings frequently emerged immediately after the IA-CC or OA-SR segments, hinting at a habitual filler strategy rather than substantive reasoning. In several cases, the noise looped, yielding near-duplicate sentences three or more times in succession. Such prolixity risks obscuring authentic evaluative logic and inflating token cost.

\subsubsection{Theme 3: Disciplinary Signatures in Reflective Patterns}
The distribution of codes varied systematically by subject area. Mathematics \& Science tasks exhibited abundant IA-P (planning) and IA-C (clarification) moves, mirroring the stepwise procedural reasoning required for the proof or calculation. History \& Education prompts elicited more IA-CC (rubric comparison) and OA-SR (holistic evaluation), congruent with discursive reconstruction and policy interpretation typical of these domains. This disciplinary fingerprint aligns with cognitive task demands: procedural subjects privilege iterative plan-execute cycles, while interpretive subjects favor comparative judgment and narrative synthesis.

\subsubsection{Theme 3: Disciplinary Signatures in Reflective Patterns}
The distribution of codes varied systematically by subject area. Mathematics \& Science tasks exhibited abundant IA-P (planning) and IA-C (clarification) moves, mirroring the stepwise procedural reasoning required for the proof or calculation. History \& Education prompts elicited more IA-CC (rubric comparison) and OA-SR (holistic evaluation), congruent with discursive reconstruction and policy interpretation typical of these domains. This disciplinary fingerprint aligns with cognitive task demands: procedural subjects privilege iterative plan-execute cycles, while interpretive subjects favor comparative judgment and narrative synthesis. Recognizing these domain-specific patterns provides a foundation for tailoring reflective scaffolds that align the model’s reasoning even more closely with expert professional practice.


\subsection{Quantified Qualitative Analysis}

\begin{figure*}
    \centering
    \includegraphics[width=1\linewidth]{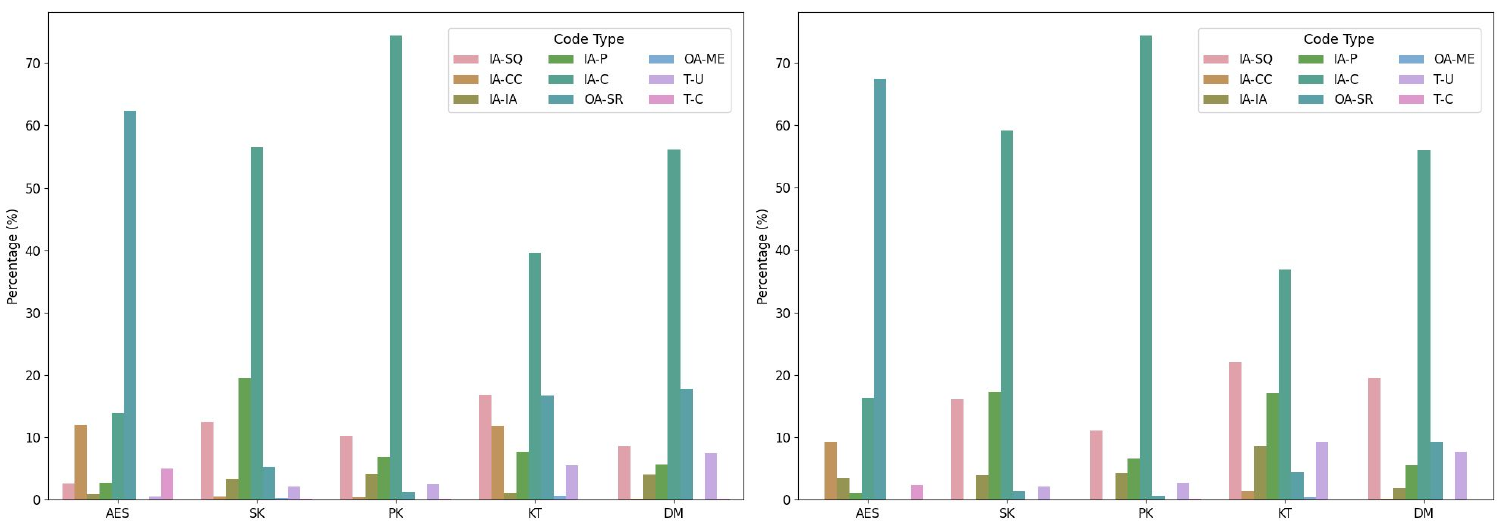}
    \captionsetup{skip=1pt}
    \caption{Quantified qualitative analyses of domain-specific data. \textit{Left} shows code type ratios by domain, and \textit{Right} shows code type ratios with COP prompting.}
    \label{fig:analysis_1.2_fig}
\end{figure*}

Using the previously developed codebook (Table \ref{tab:evaluation_codes}), we automatically coded all data with GPT-4.1 \cite{openai2025gpt41}, rather than relying solely on manual qualitative methods. This approach was necessary due to the large volume of data, which made traditional qualitative coding impractical for the entire dataset \cite{roberts2024artificial, dai2023llm, hayes2025conversing}. Note that a comprehensive machine–human agreement analysis is left for future work.

Figure \ref{fig:analysis_1.2_fig} presents a quantified qualitative analysis with each bar representing the percentage distribution of coded thinking tokens across five educational domains. The \textit{Left} shows the baseline prompting condition, while the \textit{Right} illustrates results under the CoP prompting.

In the \textit{Left}, OA-SR is overwhelmingly dominant across all domains. This is especially true in PK, where OA-SR exceeds 70\%, and remains above 50\% in AES, SK, and DM. Other codes such as IA-SQ, IA-P, and IA-CC appear with moderate frequency, particularly in SK and KT. Meanwhile, high-level metacognitive codes like OA-ME are rarely observed in any domain, and DM is characterized by slightly higher proportions of uncertainty (T-U) and conflict (T-C) codes.

In the \textit{Right}, which applies CoP prompting, the distribution of thinking codes becomes more diversified. OA-SR remains the most frequent code in all domains, but the relative proportion decreases in several areas, most notably in KT and SK. Importantly, the proportion of IA-SQ and IA-P increases in these domains, indicating that CoP prompting encourages the model to engage more in self-questioning and planning, rather than relying solely on holistic evaluation. In KT, for example, the increase in IA-SQ and IA-P suggests that the model is now more actively simulating the process of knowledge tracking with procedural and self-regulatory reasoning, thereby better reflecting the forms of reflective practice that are emphasized as professional ideals in teacher education and research.

Overall, these results suggest that CoP prompting enhances the pedagogical alignment of model reasoning by diversifying the types of cognitive processes invoked.

\section{Related work}

\subsection{Large Reasoning Models}
The release of OpenAI’s o1 \cite{jaech2024openai} marked the beginning of a new era in LRMs, transitioning from simple generation tasks to performing complex, multi-step reasoning, known as Test-time Scaling (TTS). Following this, several efforts attempted to replicate o1’s capabilities \cite{sky_t1_2025, Slow_Thinking_with_LLMs_1, Slow_Thinking_with_LLMs_2, Slow_Thinking_with_LLMs_3}. Among these, DeepSeek-R1 \cite{guo2025deepseek} demonstrated reasoning capabilities on par with o1, particularly in math and logic, thereby validating the feasibility of constructing high-performing LRMs—making it easier for researchers to replicate and build upon these models.

While LRMs demonstrate strong performance in well-structured domains such as mathematics and programming \cite{chen2025towards, guo2025deepseek}, they continue to struggle with open-ended tasks where reasoning pathways are less deterministic \cite{kim2025limitations, xu2025towards, li2025system}. Although recent studies have extended LRM applications to complex domains such as medicine and finance, these efforts have revealed persistent limitations in generalizability, reasoning depth, and adaptability to open-ended problems \cite{liu2025fin, zhang2025med, pan2025medvlm, lai2025med}. Notably, the education domain remains largely underexplored in the context of LRM-based reasoning.

To overcome this limitation, the present study proposes a pedagogical reasoning framework for educational use of LRMs, and systematically examines whether the limitations observed in domains like medicine and finance \cite{liu2025fin, zhang2025med, pan2025medvlm, lai2025med} are similarly manifested in educational contexts.

\subsection{Pedagogical Alignment and Benchmark for Education}
Pedagogical alignment is a methodology for aligning LLMs with educational objectives that involve breaking complex problems into manageable steps and providing hints and scaffolded guidance rather than direct answers \cite{sonkar2024pedagogicalalignment}. A similar concept, pedagogical steering, addresses shifting LLMs from optimizing for user satisfaction toward promoting students' cognitive engagement through hints and guiding questions in tutoring interactions \cite{puech2025pedagogicalsteering}. Both concepts emphasize effective learning outcomes over immediate user satisfaction. Pedagogical alignment encompasses two distinct approaches: general and domain-specific. Google's LearnLM-Tutor exemplifies the general approach, enhancing Gemini's pedagogical capabilities by translating learning science principles into educational benchmarks and developing specialized training datasets that educators prefer \cite{jurenka2024responsibledevelopmentgenerativeai}. Domain-specific approach includes targeted research in contexts such as art education \cite{Lee_2024} and science education \cite{chevalier2024languagemodelssciencetutors}. However, existing pedagogical alignment methods typically lack transparency in their reasoning processes, as they employ SFT \cite{sonkar2024pedagogicalalignment} or RLHF with reward models \cite{jurenka2024responsibledevelopmentgenerativeai} without explicitly revealing the rationale behind pedagogical judgments, limiting their effectiveness in complex educational situations.

Education benchmarks for evaluating LLMs' pedagogical performance have also emerged in two categories: general and domain-specific frameworks. The Cross-Domain Pedagogical Knowledge (CDPK) benchmark assesses general capabilities through teacher training examination questions spanning various subjects and educational levels, but is not openly accessible to the broader research community \cite{aiforeducationorgbench}. In contrast, MathTutorBench offers an open-source framework for domain-specific evaluation in mathematics, utilizing datasets and metrics grounded in learning sciences research \cite{macina2025mathtutorbench}. Despite these developments, comprehensive pedagogical benchmarks remain scarce in the field of Artificial Intelligence in Education (AIED). This research addresses this gap by introducing an open-source dataset that applies pedagogical reasoning, providing benchmarks with logical rationale that explicitly supports pedagogical judgments, making a significant contribution to standardizing evaluation methods in AIED.

\subsection{Reflection of Human Teacher}

Reflective practice is a foundational component of professional teaching. Human teachers engage in ongoing reflection-in-action, making adjustments to their instructional strategies during lessons, and reflection-on-action, reviewing and analyzing their teaching afterward to inform future improvements \cite{schon2017reflective, loughran2002developing}. These reflective processes not only support teachers’ immediate responses to classroom dynamics but also facilitate long-term professional growth and the development of pedagogical expertise \cite{griffiths1992using, slade2019impact}. Recent studies emphasize the importance of structured reflection for both pre-service and in-service teachers, highlighting how practices such as peer feedback, self-assessment, and the use of classroom analytics tools contribute to deeper sense-making and instructional improvement \cite{alzoubi2022data, erdemir2021reflective}. By actively reflecting on both their actions and student outcomes, teachers refine their practice, adapt to diverse learning needs, and sustain effective teaching over time. However, existing methodologies remain limited in capturing the underlying meaning of reflective practice, often reducing it to surface-level counts of reflective utterances without probing the cognitive pathways that link observation to pedagogical action \cite{treacy2021promoting, mcgarr2021use, colomer2020reflective}. Consequently, it is still unclear how these micro-steps of reflection can be operationalized and scaled within AI-supported teaching tools. To address this gap, our study analyzes the reasoning process of LRMs using a grounded theory approach \cite{corbin2017grounded}, thereby providing the  systematic account of how artificial agents can both emulate and extend human reflective cycles and offering concrete design implications for next-generation teacher-support systems.

\section{Conclusion}

This study introduced Pedagogy-R1, a pedagogical reasoning framework that adapts LLMs for educational applications through distillation-based instruction tuning and the use of pedagogically filtered outputs. By constructing the WBEB and proposing the CoP prompting strategy, we systematically evaluated model performance across key pedagogical tasks. Our results demonstrate that pedagogically aligned instruction tuning significantly improves educational reasoning, feedback, and decision-making compared to standard instruction tuning or general-purpose models. These findings highlight the importance of domain-specific alignment for deploying LLMs in authentic educational settings.

\section{GenAI Usage Disclosure}

To improve the readability of this work, generative AI models (GPT-4o, GPT-4.1) were utilized for language editing. In addition, Claude-3.7-sonnet and Gemini-2.5-Pro provided assistance during code-related tasks. All other aspects of the study—including logical argumentation, literature review, and experimental design—were conducted directly by the human authors.



@misc{colomer2020reflective,
  title={Reflective learning in higher education: Active methodologies for transformative practices},
  author={Colomer, Jordi and Serra, Teresa and Ca{\~n}abate, Dolors and Bubnys, Remigijus},
  journal={Sustainability},
  volume={12},
  number={9},
  pages={3827},
  year={2020},
  publisher={MDPI}
}

@article{mcgarr2021use,
  title={The use of virtual simulations in teacher education to develop pre-service teachers’ behaviour and classroom management skills: implications for reflective practice},
  author={McGarr, Oliver},
  journal={Journal of Education for Teaching},
  volume={47},
  number={2},
  pages={274--286},
  year={2021},
  publisher={Taylor \& Francis}
}

@article{treacy2021promoting,
  title={Promoting interconnections between reflective practice and collective creativity in higher arts education: the potential of engaging with a reflective matrix},
  author={Treacy, Danielle and Gaunt, Helena},
  journal={Reflective Practice},
  volume={22},
  number={4},
  pages={488--500},
  year={2021},
  publisher={Taylor \& Francis}
}


@article{schon2017reflective,
  title={The reflective practitioner: How professionals think in action},
  author={Sch{\"o}n, Donald A},
  year={2017},
  publisher={Routledge}
}

@article{guo2025deepseek,
  title={Deepseek-r1: Incentivizing reasoning capability in llms via reinforcement learning},
  author={Guo, Daya and Yang, Dejian and Zhang, Haowei and Song, Junxiao and Zhang, Ruoyu and Xu, Runxin and Zhu, Qihao and Ma, Shirong and Wang, Peiyi and Bi, Xiao and others},
  journal={arXiv preprint arXiv:2501.12948},
  year={2025}
}

@article{li2025mediating,
  title={The mediating effects of needs satisfaction on the relationship between teacher support and student engagement with generative artificial intelligence (GenAI) chatbots from a self-determination theory (SDT) perspective},
  author={Li, Yan and Chiu, Thomas KF},
  journal={Education and Information Technologies},
  pages={1--20},
  year={2025},
  publisher={Springer}
}

@article{zhang2024cjeval,
      title={CJEval: A Benchmark for Assessing Large Language Models Using Chinese Junior High School Exam Data}, 
      author={Qian-Wen Zhang and Haochen Wang and Fang Li and Siyu An and Lingfeng Qiao and Liangcai Gao and Di Yin and Xing Sun},
      year={2024},
      eprint={2409.16202},
      archivePrefix={arXiv},
      primaryClass={cs.AI},
}

@article{abdelrahman2022dbe,
  title={DBE-KT22: A Knowledge Tracing Dataset Based on Online Student Evaluation},
  author={Abdelrahman, Ghodai and Abdelfattah, Sherif and Wang, Qing and Lin, Yu},
  journal={arXiv preprint arXiv:2208.12651},
  year={2022}
}

@article{liu2024xes3g5m,
  title={XES3G5M: A Knowledge Tracing Benchmark Dataset with Auxiliary Information},
  author={Liu, Zitao and Liu, Qiongqiong and Guo, Teng and Chen, Jiahao and Huang, Shuyan and Zhao, Xiangyu and Tang, Jiliang and Luo, Weiqi and Weng, Jian},
  journal={Advances in Neural Information Processing Systems},
  volume={36},
  year={2024}
}

@misc{CSEDM2019,
  title = {2nd Educational Data Mining in Computer Science Education (CSEDM) Workshop},
  author = {{CSEDM Workshop}},
  year = {2019},
  month = {March},
  day = {5},
  note = {In conjunction with LAK 2019 at Arizona State University, Tempe AZ, USA},
  url = {https://sites.google.com/asu.edu/csedm-ws-lak-2019},
  howpublished = {\url{https://sites.google.com/asu.edu/csedm-ws-lak-2019}}
}

@misc{crossley2024aes,
  author       = {Scott Crossley and Perpetual Baffour and Jules King and Lauryn Burleigh and Walter Reade and Maggie Demkin},
  title        = {Learning Agency Lab - Automated Essay Scoring 2.0},
  year         = {2024},
  howpublished = {\url{https://kaggle.com/competitions/learning-agency-lab-automated-essay-scoring-2}},
  note         = {Kaggle Competition}
}

@inproceedings{demszky2023ncte,
  author    = {Dorottya Demszky and Heather Hill},
  title     = {The NCTE Transcripts: A Dataset of Elementary Math Classroom Transcripts},
  booktitle = {Proceedings of the 18th Workshop on Innovative Use of NLP for Building Educational Applications (BEA 2023)},
  year      = {2023},
  month     = {July},
  pages     = {528--538}
}

@misc{sky_t1_2025,
  author       = {NovaSky Team},
  title        = {Sky-T1: Train your own O1 preview model within \$450},
  howpublished = {https://novasky-ai.github.io/posts/sky-t1},
  note         = {Accessed: 2025-01-09},
  year         = {2025}
}

@article{Slow_Thinking_with_LLMs_1,
  title={Enhancing LLM Reasoning with Reward-guided Tree Search},
  author={Jiang, Jinhao and Chen, Zhipeng and Min, Yingqian and Chen, Jie and Cheng, Xiaoxue and Wang, Jiapeng and Tang, Yiru and Sun, Haoxiang and Deng, Jia and Zhao, Wayne Xin and Liu, Zheng and Yan, Dong and Xie, Jian and Wang, Zhongyuan and Wen, Ji-Rong},
  journal={arXiv preprint arXiv:2411.11694},
  year={2024}
}

@article{Slow_Thinking_with_LLMs_2,
  title={Imitate, Explore, and Self-Improve: A Reproduction Report on Slow-thinking Reasoning Systems},
  author={Min, Yingqian and Chen, Zhipeng and Jiang, Jinhao and Chen, Jie and Deng, Jia and Hu, Yiwen and Tang, Yiru and Wang, Jiapeng and Cheng, Xiaoxue and Song, Huatong and Zhao, Wayne Xin and Liu, Zheng and Wang, Zhongyuan and Wen, Ji-Rong},
  journal={arXiv preprint arXiv:2412.09413},
  year={2024}
}

@article{Slow_Thinking_with_LLMs_3,
      title={An Empirical Study on Eliciting and Improving R1-like Reasoning Models},
      author={Chen, Zhipeng and Min, Yingqian and Zhang, Beichen  and Chen, Jie and Jiang, Jinhao and Cheng, Daixuan and Zhao, Wayne Xin and Liu, Zheng and Miao, Xu and Lu, Yang and Fang, Lei and Wang, Zhongyuan and Wen, Ji-Rong},
      journal={arXiv preprint arXiv:2503.04548},
      year={2025}
}

@article{jaech2024openai,
  title={Openai o1 system card},
  author={Jaech, Aaron and Kalai, Adam and Lerer, Adam and Richardson, Adam and El-Kishky, Ahmed and Low, Aiden and Helyar, Alec and Madry, Aleksander and Beutel, Alex and Carney, Alex and others},
  journal={arXiv preprint arXiv:2412.16720},
  year={2024}
}

@article{chen2025towards,
  title={Towards reasoning era: A survey of long chain-of-thought for reasoning large language models},
  author={Chen, Qiguang and Qin, Libo and Liu, Jinhao and Peng, Dengyun and Guan, Jiannan and Wang, Peng and Hu, Mengkang and Zhou, Yuhang and Gao, Te and Che, Wanxiang},
  journal={arXiv preprint arXiv:2503.09567},
  year={2025}
}

@article{kim2025limitations,
  title={Limitations of Large Language Models in Clinical Problem-Solving Arising from Inflexible Reasoning},
  author={Kim, Jonathan and Podlasek, Anna and Shidara, Kie and Liu, Feng and Alaa, Ahmed and Bernardo, Danilo},
  journal={arXiv preprint arXiv:2502.04381},
  year={2025}
}

@article{xu2025towards,
  title={Towards Large Reasoning Models: A Survey of Reinforced Reasoning with Large Language Models},
  author={Xu, Fengli and Hao, Qianyue and Zong, Zefang and Wang, Jingwei and Zhang, Yunke and Wang, Jingyi and Lan, Xiaochong and Gong, Jiahui and Ouyang, Tianjian and Meng, Fanjin and others},
  journal={arXiv preprint arXiv:2501.09686},
  year={2025}
}

@article{li2025system,
  title={From system 1 to system 2: A survey of reasoning large language models},
  author={Li, Zhong-Zhi and Zhang, Duzhen and Zhang, Ming-Liang and Zhang, Jiaxin and Liu, Zengyan and Yao, Yuxuan and Xu, Haotian and Zheng, Junhao and Wang, Pei-Jie and Chen, Xiuyi and others},
  journal={arXiv preprint arXiv:2502.17419},
  year={2025}
}

@article{liu2025fin,
  title={Fin-r1: A large language model for financial reasoning through reinforcement learning},
  author={Liu, Zhaowei and Guo, Xin and Lou, Fangqi and Zeng, Lingfeng and Niu, Jinyi and Wang, Zixuan and Xu, Jiajie and Cai, Weige and Yang, Ziwei and Zhao, Xueqian and others},
  journal={arXiv preprint arXiv:2503.16252},
  year={2025}
}

@article{zhang2025med,
  title={Med-RLVR: Emerging Medical Reasoning from a 3B base model via reinforcement Learning},
  author={Zhang, Sheng and Liu, Qianchu and Qin, Guanghui and Naumann, Tristan and Poon, Hoifung},
  journal={arXiv preprint arXiv:2502.19655},
  year={2025}
}

@article{pan2025medvlm,
  title={Medvlm-r1: Incentivizing medical reasoning capability of vision-language models (vlms) via reinforcement learning},
  author={Pan, Jiazhen and Liu, Che and Wu, Junde and Liu, Fenglin and Zhu, Jiayuan and Li, Hongwei Bran and Chen, Chen and Ouyang, Cheng and Rueckert, Daniel},
  journal={arXiv preprint arXiv:2502.19634},
  year={2025}
}

@article{lai2025med,
  title={Med-r1: Reinforcement learning for generalizable medical reasoning in vision-language models},
  author={Lai, Yuxiang and Zhong, Jike and Li, Ming and Zhao, Shitian and Yang, Xiaofeng},
  journal={arXiv preprint arXiv:2503.13939},
  year={2025}
}

@misc{qwq32b,
    title = {QwQ-32B: Embracing the Power of Reinforcement Learning},
    url = {https://qwenlm.github.io/blog/qwq-32b/},
    author = {Qwen Team},
    month = {March},
    year = {2025}
}

@article{qwen2.5,
      title={Qwen2.5 Technical Report}, 
      author={An Yang and Baosong Yang and Beichen Zhang and Binyuan Hui and Bo Zheng and Bowen Yu and Chengyuan Li and Dayiheng Liu and Fei Huang and Haoran Wei and Huan Lin and Jian Yang and Jianhong Tu and Jianwei Zhang and Jianxin Yang and Jiaxi Yang and Jingren Zhou and Junyang Lin and Kai Dang and Keming Lu and Keqin Bao and Kexin Yang and Le Yu and Mei Li and Mingfeng Xue and Pei Zhang and Qin Zhu and Rui Men and Runji Lin and Tianhao Li and Tianyi Tang and Tingyu Xia and Xingzhang Ren and Xuancheng Ren and Yang Fan and Yang Su and Yichang Zhang and Yu Wan and Yuqiong Liu and Zeyu Cui and Zhenru Zhang and Zihan Qiu},
      journal={arXiv preprint arXiv:2412.15115},
      year={2024}
}

@article{wang2025thoughts,
  title={Thoughts Are All Over the Place: On the Underthinking of o1-Like LLMs},
  author={Wang, Yue and Liu, Qiuzhi and Xu, Jiahao and Liang, Tian and Chen, Xingyu and He, Zhiwei and Song, Linfeng and Yu, Dian and Li, Juntao and Zhang, Zhuosheng and others},
  journal={arXiv preprint arXiv:2501.18585},
  year={2025}
}

@article{chen2025seal,
  title={SEAL: Steerable Reasoning Calibration of Large Language Models for Free},
  author={Chen, Runjin and Zhang, Zhenyu and Hong, Junyuan and Kundu, Souvik and Wang, Zhangyang},
  journal={arXiv preprint arXiv:2504.07986},
  year={2025}
}

@article{sonkar2024pedagogicalalignment,
      title={Pedagogical Alignment of Large Language Models}, 
      author={Shashank Sonkar and Kangqi Ni and Sapana Chaudhary and Richard G. Baraniuk},
      year={2024},
      journal={arXiv preprint arXiv:2402.05000},
}

@article{puech2025pedagogicalsteering,
      title={Towards the Pedagogical Steering of Large Language Models for Tutoring: A Case Study with Modeling Productive Failure}, 
      author={Romain Puech and Jakub Macina and Julia Chatain and Mrinmaya Sachan and Manu Kapur},
      year={2025},
      journal={arXiv preprint arXiv:2410.03781},
}

@misc{jurenka2024responsibledevelopmentgenerativeai,
      title={Towards Responsible Development of Generative AI for Education: An Evaluation-Driven Approach}, 
      author={Irina Jurenka and Markus Kunesch and Kevin R. McKee and Daniel Gillick and Shaojian Zhu and Sara Wiltberger, ... and Lila Ibrahim},
      year={2024},
      eprint={2407.12687},
      archivePrefix={arXiv},
      primaryClass={cs.CY},
      url={https://arxiv.org/abs/2407.12687}, 
}

@article{Lee_2024,
   title={LLaVA-docent: Instruction tuning with multimodal large language model to support art appreciation education},
   volume={7},
   ISSN={2666-920X},
   url={http://dx.doi.org/10.1016/j.caeai.2024.100297},
   DOI={10.1016/j.caeai.2024.100297},
   journal={Computers and Education: Artificial Intelligence},
   publisher={Elsevier BV},
   author={Lee, Unggi and Jeon, Minji and Lee, Yunseo and Byun, Gyuri and Son, Yoorim and Shin, Jaeyoon and Ko, Hongkyu and Kim, Hyeoncheol},
   year={2024},
   month=dec, pages={100297} }

@misc{chevalier2024languagemodelssciencetutors,
      title={Language Models as Science Tutors}, 
      author={Alexis Chevalier and Jiayi Geng and Alexander Wettig and Howard Chen and Sebastian Mizera and Toni Annala, ... and Danqi Chen},
      year={2024},
      eprint={2402.11111},
      archivePrefix={arXiv},
      primaryClass={cs.CL},
      url={https://arxiv.org/abs/2402.11111}, 
}

@article{jung2024clst,
  title={CLST: Cold-Start Mitigation in Knowledge Tracing by Aligning a Generative Language Model as a Students' Knowledge Tracer},
  author={Jung, Heeseok and Yoo, Jaesang and Yoon, Yohaan and Jang, Yeonju},
  journal={arXiv preprint arXiv:2406.10296},
  year={2024}
}

@article{lee2024language,
  title={Language model can do knowledge tracing: Simple but effective method to integrate language model and knowledge tracing task},
  author={Lee, Unggi and Bae, Jiyeong and Kim, Dohee and Lee, Sookbun and Park, Jaekwon and Ahn, Taekyung and Lee, Gunho and Stratton, Damji and Kim, Hyeoncheol},
  journal={arXiv preprint arXiv:2406.02893},
  year={2024}
}

@inproceedings{zhan2024knowledge,
  title={Knowledge tracing as language processing: a large-scale autoregressive paradigm},
  author={Zhan, Bojun and Guo, Teng and Li, Xueyi and Hou, Mingliang and Liang, Qianru and Gao, Boyu and Luo, Weiqi and Liu, Zitao},
  booktitle={International Conference on Artificial Intelligence in Education},
  pages={177--191},
  year={2024},
  organization={Springer}
}

@inproceedings{lee2024difficulty,
  title={Difficulty-Focused Contrastive Learning for Knowledge Tracing with a Large Language Model-Based Difficulty Prediction},
  author={Lee, Unggi and Yoon, Sungjun and Yun, Joon Seo and Park, Kyoungsoo and Jung, Young Hoon and Stratton, Damji and Kim, Hyeoncheol},
  booktitle={Joint 30th International Conference on Computational Linguistics and 14th International Conference on Language Resources and Evaluation, LREC-COLING 2024},
  pages={4891--4900},
  year={2024},
  organization={European Language Resources Association (ELRA)}
}

@online{openai2025gpt41,
  author       = {OpenAI},
  title        = {Introducing GPT-4.1 in the API},
  year         = {2025},
  month        = apr,
  url          = {https://openai.com/index/gpt-4-1/},
  note         = {Accessed: 2025-05-17}
}


@misc{aiforeducationorgbench,
    title = {The Pedagogy Benchmark},
    url = {https://benchmarks.ai-for-education.org/#moreinfo-about-the-benchmark},
    author = {AI-For-Education.org},
    month = {May},
    year = {2025}
}


@article{macina2025mathtutorbench,
      title={MathTutorBench: A Benchmark for Measuring Open-ended Pedagogical Capabilities of LLM Tutors}, 
      author={Jakub Macina and Nico Daheim and Ido Hakimi and Manu Kapur and Iryna Gurevych and Mrinmaya Sachan},
      year={2025},
      journal={arXiv preprint arXiv:2502.18940},
}

@article{griot2025large,
  title={Large language models lack essential metacognition for reliable medical reasoning},
  author={Griot, Maxime and Hemptinne, Coralie and Vanderdonckt, Jean and Yuksel, Demet},
  journal={Nature communications},
  volume={16},
  number={1},
  pages={642},
  year={2025},
  publisher={Nature Publishing Group UK London}
}

@techreport{jurenka2024learnlm,
  title     = {Towards Responsible Development of Generative AI for Education: An Evaluation-Driven Approach},
  author    = {Irina Jurenka and Markus Kunesch and Kevin R. McKee and Daniel Gillick and Shaojian Zhu and Sara Wiltberger and Shubham Milind Phal and Katherine Hermann and Daniel Kasenberg and Avishkar Bhoopchand and Ankit Anand and Miruna Pîslar and Stephanie Chan and Lisa Wang and Jennifer She and Parsa Mahmoudieh and Aliya Rysbek and Wei-Jen Ko and Andrea Huber and Brett Wiltshire and Gal Elidan and Roni Rabin and Jasmin Rubinovitz and Amit Pitaru and Mac McAllister and Julia Wilkowski and David Choi and Roee Engelberg and Lidan Hackmon and Adva Levin and Rachel Griffin and Michael Sears and Filip Bar and Mia Mesar and Mana Jabbour and Arslan Chaudhry and James Cohan and Sridhar Thiagarajan and Nir Levine and Ben Brown and Dilan Gorur and Svetlana Grant and Rachel Hashimshoni and Laura Weidinger and Jieru Hu and Dawn Chen and Kuba Dolecki and Canfer Akbulut and Maxwell Bileschi and Laura Culp and Wen-Xin Dong and Nahema Marchal and Kelsie Van Deman and Hema Bajaj Misra and Michael Duah and Moran Ambar and Avi Caciularu and Sandra Lefdal and Chris Summerfield and James An and Pierre-Alexandre Kamienny and Abhinit Mohdi and Theofilos Strinopoulous and Annie Hale and Wayne Anderson and Luis C. Cobo and Niv Efron and Muktha Ananda and Shakir Mohamed and Maureen Heymans and Zoubin Ghahramani and Yossi Matias and Ben Gomes and Lila Ibrahim},
  institution = {Google DeepMind},
  year      = {2024},
  url       = {https://goo.gle/LearnLM},
  note      = {Technical Report}
}

@article{kasneci2023chatgpt,
  title={ChatGPT for good? On opportunities and challenges of large language models for education},
  author={Kasneci, Enkelejda and Se{\ss}ler, Kathrin and K{\"u}chemann, Stefan and Bannert, Maria and Dementieva, Daryna and Fischer, Frank and Gasser, Urs and Groh, Georg and G{\"u}nnemann, Stephan and H{\"u}llermeier, Eyke and others},
  journal={Learning and individual differences},
  volume={103},
  pages={102274},
  year={2023},
  publisher={Elsevier}
}

@article{yan2024practical,
  title={Practical and ethical challenges of large language models in education: A systematic scoping review},
  author={Yan, Lixiang and Sha, Lele and Zhao, Linxuan and Li, Yuheng and Martinez-Maldonado, Roberto and Chen, Guanliang and Li, Xinyu and Jin, Yueqiao and Ga{\v{s}}evi{\'c}, Dragan},
  journal={British Journal of Educational Technology},
  volume={55},
  number={1},
  pages={90--112},
  year={2024},
  publisher={Wiley Online Library}
}

@article{wang2024large,
  title={Large language models for education: A survey and outlook},
  author={Wang, Shen and Xu, Tianlong and Li, Hang and Zhang, Chaoli and Liang, Joleen and Tang, Jiliang and Yu, Philip S and Wen, Qingsong},
  journal={arXiv preprint arXiv:2403.18105},
  year={2024}
}

@article{jeon2023large,
  title={Large language models in education: A focus on the complementary relationship between human teachers and ChatGPT},
  author={Jeon, Jaeho and Lee, Seongyong},
  journal={Education and Information Technologies},
  volume={28},
  number={12},
  pages={15873--15892},
  year={2023},
  publisher={Springer}
}

@book{glaser2017discovery,
  title={Discovery of grounded theory: Strategies for qualitative research},
  author={Glaser, Barney and Strauss, Anselm},
  year={2017},
  publisher={Routledge}
}

@article{baidoo2023education,
  title={Education in the era of generative artificial intelligence (AI): Understanding the potential benefits of ChatGPT in promoting teaching and learning},
  author={Baidoo-Anu, David and Ansah, Leticia Owusu},
  journal={Journal of AI},
  volume={7},
  number={1},
  pages={52--62},
  year={2023},
  publisher={{\.I}zmir Academy Association}
}

@article{whalen2023chatgpt,
  title={ChatGPT: Challenges, opportunities, and implications for teacher education},
  author={Whalen, Jeromie and Mouza, Chrystalla and others},
  journal={Contemporary Issues in Technology and Teacher Education},
  volume={23},
  number={1},
  pages={1--23},
  year={2023},
  publisher={Society for Information Technology \& Teacher Education}
}

@article{rasul2023role,
  title={The role of ChatGPT in higher education: Benefits, challenges, and future research directions},
  author={Rasul, Tareq and Nair, Sumesh and Kalendra, Diane and Robin, Mulyadi and de Oliveira Santini, Fernando and Ladeira, Wagner Junior and Sun, Mingwei and Day, Ingrid and Rather, Raouf Ahmad and Heathcote, Liz},
  journal={Journal of Applied Learning and Teaching},
  volume={6},
  number={1},
  pages={41--56},
  year={2023}
}

@incollection{borko2013teacher,
  title={Teacher decision making 1},
  author={Borko, Hilda and Shavelson, Richard J},
  booktitle={Dimensions of thinking and cognitive instruction},
  pages={311--346},
  year={2013},
  publisher={Routledge}
}

@book{eggleston2018teacher,
  title={Teacher decision-making in the classroom: A collection of papers},
  author={Eggleston, John},
  year={2018},
  publisher={Routledge}
}

@article{siuty2018unraveling,
  title={Unraveling the role of curriculum in teacher decision making},
  author={Siuty, Molly Baustien and Leko, Melinda M and Knackstedt, Kimberly M},
  journal={Teacher Education and Special Education},
  volume={41},
  number={1},
  pages={39--57},
  year={2018},
  publisher={SAGE Publications Sage CA: Los Angeles, CA}
}

@article{doyle1977practicality,
  title={The practicality ethic in teacher decision-making},
  author={Doyle, Walter and Ponder, Gerald A},
  journal={Interchange},
  volume={8},
  number={3},
  pages={1--12},
  year={1977},
  publisher={Kluwer Academic Publishers Dordrecht}
}

@article{yu2025evaluating,
  title={Evaluating test-time scaling llms for legal reasoning: Openai o1, deepseek-r1, and beyond},
  author={Yu, Yaoyao and Gan, Leilei and Hu, Yinghao and Wei, Bin and Kuang, Kun and Wu, Fei},
  journal={arXiv preprint arXiv:2503.16040},
  year={2025}
}

@article{huang2025m1,
  title={m1: Unleash the Potential of Test-Time Scaling for Medical Reasoning with Large Language Models},
  author={Huang, Xiaoke and Wu, Juncheng and Liu, Hui and Tang, Xianfeng and Zhou, Yuyin},
  journal={arXiv preprint arXiv:2504.00869},
  year={2025}
}

@article{fagbohun2024beyond,
  title={Beyond traditional assessment: Exploring the impact of large language models on grading practices},
  author={Fagbohun, Oluwole and Iduwe, Nwaamaka Pearl and Abdullahi, Mustapha and Ifaturoti, Adeseye and Nwanna, OM},
  journal={Journal of Artificial Intelligence, Machine Learning and Data Science},
  volume={2},
  number={1},
  pages={1--8},
  year={2024}
}

@article{gao2024automatic,
  title={Automatic assessment of text-based responses in post-secondary education: A systematic review},
  author={Gao, Rujun and Merzdorf, Hillary E and Anwar, Saira and Hipwell, M Cynthia and Srinivasa, Arun R},
  journal={Computers and Education: Artificial Intelligence},
  volume={6},
  pages={100206},
  year={2024},
  publisher={Elsevier}
}

@article{mizumoto2023exploring,
  title={Exploring the potential of using an AI language model for automated essay scoring},
  author={Mizumoto, Atsushi and Eguchi, Masaki},
  journal={Research Methods in Applied Linguistics},
  volume={2},
  number={2},
  pages={100050},
  year={2023},
  publisher={Elsevier}
}

@inproceedings{fu2024sinkt,
  title={Sinkt: A structure-aware inductive knowledge tracing model with large language model},
  author={Fu, Lingyue and Guan, Hao and Du, Kounianhua and Lin, Jianghao and Xia, Wei and Zhang, Weinan and Tang, Ruiming and Wang, Yasheng and Yu, Yong},
  booktitle={Proceedings of the 33rd ACM International Conference on Information and Knowledge Management},
  pages={632--642},
  year={2024}
}

@article{glaser1998grounded,
  title={Grounded theory},
  author={Glaser, Barney G and Strauss, Anselm L},
  journal={Strategien qualitativer Forschung. Bern: Huber},
  volume={4},
  year={1998},
  publisher={Springer}
}

@article{corbin2017grounded,
  title={Grounded theory},
  author={Corbin, Juliet},
  journal={The Journal of Positive Psychology},
  volume={12},
  number={3},
  pages={301--302},
  year={2017},
  publisher={Taylor \& Francis}
}

@article{roberts2024artificial,
  title={Artificial intelligence and qualitative research: The promise and perils of large language model (LLM)‘assistance’},
  author={Roberts, John and Baker, Max and Andrew, Jane},
  journal={Critical Perspectives on Accounting},
  volume={99},
  pages={102722},
  year={2024},
  publisher={Elsevier}
}

@inproceedings{dai2023llm,
  title={LLM-in-the-loop: Leveraging Large Language Model for Thematic Analysis},
  author={Dai, Shih-Chieh and Xiong, Aiping and Ku, Lun-Wei},
  booktitle={Findings of the Association for Computational Linguistics: EMNLP 2023},
  pages={9993--10001},
  year={2023}
}

@article{hayes2025conversing,
  title={“Conversing” With Qualitative Data: Enhancing Qualitative Research Through Large Language Models (LLMs)},
  author={Hayes, Adam S},
  journal={International Journal of Qualitative Methods},
  volume={24},
  pages={16094069251322346},
  year={2025},
  publisher={SAGE Publications Sage CA: Los Angeles, CA}
}

@book{loughran2002developing,
  title={Developing reflective practice: Learning about teaching and learning through modelling},
  author={Loughran, J John},
  year={2002},
  publisher={Routledge}
}

@article{griffiths1992using,
  title={Using reflective practice to link personal and public theories},
  author={Griffiths, Morwenna and Tann, Sarah},
  journal={Journal of Education for teaching},
  volume={18},
  number={1},
  pages={69--84},
  year={1992},
  publisher={Taylor \& Francis}
}

@article{slade2019impact,
  title={The Impact of Reflective Practice on Teacher Candidates' Learning.},
  author={Slade, Mary L and Burnham, Tammy J and Catalana, Sarah Marie and Waters, Tammy},
  journal={International Journal for the Scholarship of Teaching and Learning},
  volume={13},
  number={2},
  pages={15},
  year={2019},
  publisher={ERIC}
}

@inproceedings{alzoubi2022data,
  title={From data to actions: Unfolding instructors’ sense-making and reflective practice with classroom analytics},
  author={AlZoubi, Dana},
  booktitle={Proceedings of 12th International Conference on Learning Analytics and Knowledge (LAK22)},
  year={2022}
}

@article{erdemir2021reflective,
  title={Reflective practices in micro teaching from the perspective of preservice teachers: Teacher feedback, peer feedback and self-reflection},
  author={Erdemir, Nihan and Ye{\c{s}}il{\c{c}}{\i}nar, Sabahattin},
  journal={Reflective Practice},
  volume={22},
  number={6},
  pages={766--781},
  year={2021},
  publisher={Taylor \& Francis}
}

@article{nelson2020computational,
  title={Computational grounded theory: A methodological framework},
  author={Nelson, Laura K},
  journal={Sociological Methods \& Research},
  volume={49},
  number={1},
  pages={3--42},
  year={2020},
  publisher={Sage Publications Sage CA: Los Angeles, CA}
}

\begin{thebibliography}{71}


\ifx \showCODEN    \undefined \def \showCODEN     #1{\unskip}     \fi
\ifx \showISBNx    \undefined \def \showISBNx     #1{\unskip}     \fi
\ifx \showISBNxiii \undefined \def \showISBNxiii  #1{\unskip}     \fi
\ifx \showISSN     \undefined \def \showISSN      #1{\unskip}     \fi
\ifx \showLCCN     \undefined \def \showLCCN      #1{\unskip}     \fi
\ifx \shownote     \undefined \def \shownote      #1{#1}          \fi
\ifx \showarticletitle \undefined \def \showarticletitle #1{#1}   \fi
\ifx \showURL      \undefined \def \showURL       {\relax}        \fi
\providecommand\bibfield[2]{#2}
\providecommand\bibinfo[2]{#2}
\providecommand\natexlab[1]{#1}
\providecommand\showeprint[2][]{arXiv:#2}

\bibitem[Abdelrahman et~al\mbox{.}(2022)]%
        {abdelrahman2022dbe}
\bibfield{author}{\bibinfo{person}{Ghodai Abdelrahman}, \bibinfo{person}{Sherif Abdelfattah}, \bibinfo{person}{Qing Wang}, {and} \bibinfo{person}{Yu Lin}.} \bibinfo{year}{2022}\natexlab{}.
\newblock \showarticletitle{DBE-KT22: A Knowledge Tracing Dataset Based on Online Student Evaluation}.
\newblock \bibinfo{journal}{\emph{arXiv preprint arXiv:2208.12651}} (\bibinfo{year}{2022}).
\newblock


\bibitem[AI-For-Education.org(2025)]%
        {aiforeducationorgbench}
\bibfield{author}{\bibinfo{person}{AI-For-Education.org}.} \bibinfo{year}{2025}\natexlab{}.
\newblock \bibinfo{title}{The Pedagogy Benchmark}.
\newblock
\urldef\tempurl%
\url{https://benchmarks.ai-for-education.org/#moreinfo-about-the-benchmark}
\showURL{%
\tempurl}


\bibitem[AlZoubi(2022)]%
        {alzoubi2022data}
\bibfield{author}{\bibinfo{person}{Dana AlZoubi}.} \bibinfo{year}{2022}\natexlab{}.
\newblock \showarticletitle{From data to actions: Unfolding instructors’ sense-making and reflective practice with classroom analytics}. In \bibinfo{booktitle}{\emph{Proceedings of 12th International Conference on Learning Analytics and Knowledge (LAK22)}}.
\newblock


\bibitem[Baidoo-Anu and Ansah(2023)]%
        {baidoo2023education}
\bibfield{author}{\bibinfo{person}{David Baidoo-Anu} {and} \bibinfo{person}{Leticia~Owusu Ansah}.} \bibinfo{year}{2023}\natexlab{}.
\newblock \showarticletitle{Education in the era of generative artificial intelligence (AI): Understanding the potential benefits of ChatGPT in promoting teaching and learning}.
\newblock \bibinfo{journal}{\emph{Journal of AI}} \bibinfo{volume}{7}, \bibinfo{number}{1} (\bibinfo{year}{2023}), \bibinfo{pages}{52--62}.
\newblock


\bibitem[Borko and Shavelson(2013)]%
        {borko2013teacher}
\bibfield{author}{\bibinfo{person}{Hilda Borko} {and} \bibinfo{person}{Richard~J Shavelson}.} \bibinfo{year}{2013}\natexlab{}.
\newblock \showarticletitle{Teacher decision making 1}.
\newblock In \bibinfo{booktitle}{\emph{Dimensions of thinking and cognitive instruction}}. \bibinfo{publisher}{Routledge}, \bibinfo{pages}{311--346}.
\newblock


\bibitem[Chen et~al\mbox{.}(2025b)]%
        {chen2025towards}
\bibfield{author}{\bibinfo{person}{Qiguang Chen}, \bibinfo{person}{Libo Qin}, \bibinfo{person}{Jinhao Liu}, \bibinfo{person}{Dengyun Peng}, \bibinfo{person}{Jiannan Guan}, \bibinfo{person}{Peng Wang}, \bibinfo{person}{Mengkang Hu}, \bibinfo{person}{Yuhang Zhou}, \bibinfo{person}{Te Gao}, {and} \bibinfo{person}{Wanxiang Che}.} \bibinfo{year}{2025}\natexlab{b}.
\newblock \showarticletitle{Towards reasoning era: A survey of long chain-of-thought for reasoning large language models}.
\newblock \bibinfo{journal}{\emph{arXiv preprint arXiv:2503.09567}} (\bibinfo{year}{2025}).
\newblock


\bibitem[Chen et~al\mbox{.}(2025c)]%
        {chen2025seal}
\bibfield{author}{\bibinfo{person}{Runjin Chen}, \bibinfo{person}{Zhenyu Zhang}, \bibinfo{person}{Junyuan Hong}, \bibinfo{person}{Souvik Kundu}, {and} \bibinfo{person}{Zhangyang Wang}.} \bibinfo{year}{2025}\natexlab{c}.
\newblock \showarticletitle{SEAL: Steerable Reasoning Calibration of Large Language Models for Free}.
\newblock \bibinfo{journal}{\emph{arXiv preprint arXiv:2504.07986}} (\bibinfo{year}{2025}).
\newblock


\bibitem[Chen et~al\mbox{.}(2025a)]%
        {Slow_Thinking_with_LLMs_3}
\bibfield{author}{\bibinfo{person}{Zhipeng Chen}, \bibinfo{person}{Yingqian Min}, \bibinfo{person}{Beichen Zhang}, \bibinfo{person}{Jie Chen}, \bibinfo{person}{Jinhao Jiang}, \bibinfo{person}{Daixuan Cheng}, \bibinfo{person}{Wayne~Xin Zhao}, \bibinfo{person}{Zheng Liu}, \bibinfo{person}{Xu Miao}, \bibinfo{person}{Yang Lu}, \bibinfo{person}{Lei Fang}, \bibinfo{person}{Zhongyuan Wang}, {and} \bibinfo{person}{Ji-Rong Wen}.} \bibinfo{year}{2025}\natexlab{a}.
\newblock \showarticletitle{An Empirical Study on Eliciting and Improving R1-like Reasoning Models}.
\newblock \bibinfo{journal}{\emph{arXiv preprint arXiv:2503.04548}} (\bibinfo{year}{2025}).
\newblock


\bibitem[Chevalier et~al\mbox{.}(2024)]%
        {chevalier2024languagemodelssciencetutors}
\bibfield{author}{\bibinfo{person}{Alexis Chevalier}, \bibinfo{person}{Jiayi Geng}, \bibinfo{person}{Alexander Wettig}, \bibinfo{person}{Howard Chen}, \bibinfo{person}{Sebastian Mizera}, \bibinfo{person}{... Toni~Annala}, {and} \bibinfo{person}{Danqi Chen}.} \bibinfo{year}{2024}\natexlab{}.
\newblock \bibinfo{title}{Language Models as Science Tutors}.
\newblock
\showeprint[arxiv]{2402.11111}~[cs.CL]
\urldef\tempurl%
\url{https://arxiv.org/abs/2402.11111}
\showURL{%
\tempurl}


\bibitem[Colomer et~al\mbox{.}(2020)]%
        {colomer2020reflective}
\bibfield{author}{\bibinfo{person}{Jordi Colomer}, \bibinfo{person}{Teresa Serra}, \bibinfo{person}{Dolors Ca{\~n}abate}, {and} \bibinfo{person}{Remigijus Bubnys}.} \bibinfo{year}{2020}\natexlab{}.
\newblock \bibinfo{title}{Reflective learning in higher education: Active methodologies for transformative practices}.
\newblock \bibinfo{numpages}{3827}~pages.
\newblock


\bibitem[Corbin(2017)]%
        {corbin2017grounded}
\bibfield{author}{\bibinfo{person}{Juliet Corbin}.} \bibinfo{year}{2017}\natexlab{}.
\newblock \showarticletitle{Grounded theory}.
\newblock \bibinfo{journal}{\emph{The Journal of Positive Psychology}} \bibinfo{volume}{12}, \bibinfo{number}{3} (\bibinfo{year}{2017}), \bibinfo{pages}{301--302}.
\newblock


\bibitem[Crossley et~al\mbox{.}(2024)]%
        {crossley2024aes}
\bibfield{author}{\bibinfo{person}{Scott Crossley}, \bibinfo{person}{Perpetual Baffour}, \bibinfo{person}{Jules King}, \bibinfo{person}{Lauryn Burleigh}, \bibinfo{person}{Walter Reade}, {and} \bibinfo{person}{Maggie Demkin}.} \bibinfo{year}{2024}\natexlab{}.
\newblock \bibinfo{title}{Learning Agency Lab - Automated Essay Scoring 2.0}.
\newblock \bibinfo{howpublished}{\url{https://kaggle.com/competitions/learning-agency-lab-automated-essay-scoring-2}}.
\newblock
\newblock
\shownote{Kaggle Competition}.


\bibitem[{CSEDM Workshop}(2019)]%
        {CSEDM2019}
\bibfield{author}{\bibinfo{person}{{CSEDM Workshop}}.} \bibinfo{year}{2019}\natexlab{}.
\newblock \bibinfo{title}{2nd Educational Data Mining in Computer Science Education (CSEDM) Workshop}.
\newblock \bibinfo{howpublished}{\url{https://sites.google.com/asu.edu/csedm-ws-lak-2019}}.
\newblock
\urldef\tempurl%
\url{https://sites.google.com/asu.edu/csedm-ws-lak-2019}
\showURL{%
\tempurl}
\newblock
\shownote{In conjunction with LAK 2019 at Arizona State University, Tempe AZ, USA}.


\bibitem[Dai et~al\mbox{.}(2023)]%
        {dai2023llm}
\bibfield{author}{\bibinfo{person}{Shih-Chieh Dai}, \bibinfo{person}{Aiping Xiong}, {and} \bibinfo{person}{Lun-Wei Ku}.} \bibinfo{year}{2023}\natexlab{}.
\newblock \showarticletitle{LLM-in-the-loop: Leveraging Large Language Model for Thematic Analysis}. In \bibinfo{booktitle}{\emph{Findings of the Association for Computational Linguistics: EMNLP 2023}}. \bibinfo{pages}{9993--10001}.
\newblock


\bibitem[Demszky and Hill(2023)]%
        {demszky2023ncte}
\bibfield{author}{\bibinfo{person}{Dorottya Demszky} {and} \bibinfo{person}{Heather Hill}.} \bibinfo{year}{2023}\natexlab{}.
\newblock \showarticletitle{The NCTE Transcripts: A Dataset of Elementary Math Classroom Transcripts}. In \bibinfo{booktitle}{\emph{Proceedings of the 18th Workshop on Innovative Use of NLP for Building Educational Applications (BEA 2023)}}. \bibinfo{pages}{528--538}.
\newblock


\bibitem[Doyle and Ponder(1977)]%
        {doyle1977practicality}
\bibfield{author}{\bibinfo{person}{Walter Doyle} {and} \bibinfo{person}{Gerald~A Ponder}.} \bibinfo{year}{1977}\natexlab{}.
\newblock \showarticletitle{The practicality ethic in teacher decision-making}.
\newblock \bibinfo{journal}{\emph{Interchange}} \bibinfo{volume}{8}, \bibinfo{number}{3} (\bibinfo{year}{1977}), \bibinfo{pages}{1--12}.
\newblock


\bibitem[Eggleston(2018)]%
        {eggleston2018teacher}
\bibfield{author}{\bibinfo{person}{John Eggleston}.} \bibinfo{year}{2018}\natexlab{}.
\newblock \bibinfo{booktitle}{\emph{Teacher decision-making in the classroom: A collection of papers}}.
\newblock \bibinfo{publisher}{Routledge}.
\newblock


\bibitem[Erdemir and Ye{\c{s}}il{\c{c}}{\i}nar(2021)]%
        {erdemir2021reflective}
\bibfield{author}{\bibinfo{person}{Nihan Erdemir} {and} \bibinfo{person}{Sabahattin Ye{\c{s}}il{\c{c}}{\i}nar}.} \bibinfo{year}{2021}\natexlab{}.
\newblock \showarticletitle{Reflective practices in micro teaching from the perspective of preservice teachers: Teacher feedback, peer feedback and self-reflection}.
\newblock \bibinfo{journal}{\emph{Reflective Practice}} \bibinfo{volume}{22}, \bibinfo{number}{6} (\bibinfo{year}{2021}), \bibinfo{pages}{766--781}.
\newblock


\bibitem[Fagbohun et~al\mbox{.}(2024)]%
        {fagbohun2024beyond}
\bibfield{author}{\bibinfo{person}{Oluwole Fagbohun}, \bibinfo{person}{Nwaamaka~Pearl Iduwe}, \bibinfo{person}{Mustapha Abdullahi}, \bibinfo{person}{Adeseye Ifaturoti}, {and} \bibinfo{person}{OM Nwanna}.} \bibinfo{year}{2024}\natexlab{}.
\newblock \showarticletitle{Beyond traditional assessment: Exploring the impact of large language models on grading practices}.
\newblock \bibinfo{journal}{\emph{Journal of Artificial Intelligence, Machine Learning and Data Science}} \bibinfo{volume}{2}, \bibinfo{number}{1} (\bibinfo{year}{2024}), \bibinfo{pages}{1--8}.
\newblock


\bibitem[Fu et~al\mbox{.}(2024)]%
        {fu2024sinkt}
\bibfield{author}{\bibinfo{person}{Lingyue Fu}, \bibinfo{person}{Hao Guan}, \bibinfo{person}{Kounianhua Du}, \bibinfo{person}{Jianghao Lin}, \bibinfo{person}{Wei Xia}, \bibinfo{person}{Weinan Zhang}, \bibinfo{person}{Ruiming Tang}, \bibinfo{person}{Yasheng Wang}, {and} \bibinfo{person}{Yong Yu}.} \bibinfo{year}{2024}\natexlab{}.
\newblock \showarticletitle{Sinkt: A structure-aware inductive knowledge tracing model with large language model}. In \bibinfo{booktitle}{\emph{Proceedings of the 33rd ACM International Conference on Information and Knowledge Management}}. \bibinfo{pages}{632--642}.
\newblock


\bibitem[Gao et~al\mbox{.}(2024)]%
        {gao2024automatic}
\bibfield{author}{\bibinfo{person}{Rujun Gao}, \bibinfo{person}{Hillary~E Merzdorf}, \bibinfo{person}{Saira Anwar}, \bibinfo{person}{M~Cynthia Hipwell}, {and} \bibinfo{person}{Arun~R Srinivasa}.} \bibinfo{year}{2024}\natexlab{}.
\newblock \showarticletitle{Automatic assessment of text-based responses in post-secondary education: A systematic review}.
\newblock \bibinfo{journal}{\emph{Computers and Education: Artificial Intelligence}}  \bibinfo{volume}{6} (\bibinfo{year}{2024}), \bibinfo{pages}{100206}.
\newblock


\bibitem[Glaser and Strauss(2017)]%
        {glaser2017discovery}
\bibfield{author}{\bibinfo{person}{Barney Glaser} {and} \bibinfo{person}{Anselm Strauss}.} \bibinfo{year}{2017}\natexlab{}.
\newblock \bibinfo{booktitle}{\emph{Discovery of grounded theory: Strategies for qualitative research}}.
\newblock \bibinfo{publisher}{Routledge}.
\newblock


\bibitem[Glaser and Strauss(1998)]%
        {glaser1998grounded}
\bibfield{author}{\bibinfo{person}{Barney~G Glaser} {and} \bibinfo{person}{Anselm~L Strauss}.} \bibinfo{year}{1998}\natexlab{}.
\newblock \showarticletitle{Grounded theory}.
\newblock \bibinfo{journal}{\emph{Strategien qualitativer Forschung. Bern: Huber}}  \bibinfo{volume}{4} (\bibinfo{year}{1998}).
\newblock


\bibitem[Griffiths and Tann(1992)]%
        {griffiths1992using}
\bibfield{author}{\bibinfo{person}{Morwenna Griffiths} {and} \bibinfo{person}{Sarah Tann}.} \bibinfo{year}{1992}\natexlab{}.
\newblock \showarticletitle{Using reflective practice to link personal and public theories}.
\newblock \bibinfo{journal}{\emph{Journal of Education for teaching}} \bibinfo{volume}{18}, \bibinfo{number}{1} (\bibinfo{year}{1992}), \bibinfo{pages}{69--84}.
\newblock


\bibitem[Griot et~al\mbox{.}(2025)]%
        {griot2025large}
\bibfield{author}{\bibinfo{person}{Maxime Griot}, \bibinfo{person}{Coralie Hemptinne}, \bibinfo{person}{Jean Vanderdonckt}, {and} \bibinfo{person}{Demet Yuksel}.} \bibinfo{year}{2025}\natexlab{}.
\newblock \showarticletitle{Large language models lack essential metacognition for reliable medical reasoning}.
\newblock \bibinfo{journal}{\emph{Nature communications}} \bibinfo{volume}{16}, \bibinfo{number}{1} (\bibinfo{year}{2025}), \bibinfo{pages}{642}.
\newblock


\bibitem[Guo et~al\mbox{.}(2025)]%
        {guo2025deepseek}
\bibfield{author}{\bibinfo{person}{Daya Guo}, \bibinfo{person}{Dejian Yang}, \bibinfo{person}{Haowei Zhang}, \bibinfo{person}{Junxiao Song}, \bibinfo{person}{Ruoyu Zhang}, \bibinfo{person}{Runxin Xu}, \bibinfo{person}{Qihao Zhu}, \bibinfo{person}{Shirong Ma}, \bibinfo{person}{Peiyi Wang}, \bibinfo{person}{Xiao Bi}, {et~al\mbox{.}}} \bibinfo{year}{2025}\natexlab{}.
\newblock \showarticletitle{Deepseek-r1: Incentivizing reasoning capability in llms via reinforcement learning}.
\newblock \bibinfo{journal}{\emph{arXiv preprint arXiv:2501.12948}} (\bibinfo{year}{2025}).
\newblock


\bibitem[Hayes(2025)]%
        {hayes2025conversing}
\bibfield{author}{\bibinfo{person}{Adam~S Hayes}.} \bibinfo{year}{2025}\natexlab{}.
\newblock \showarticletitle{“Conversing” With Qualitative Data: Enhancing Qualitative Research Through Large Language Models (LLMs)}.
\newblock \bibinfo{journal}{\emph{International Journal of Qualitative Methods}}  \bibinfo{volume}{24} (\bibinfo{year}{2025}), \bibinfo{pages}{16094069251322346}.
\newblock


\bibitem[Huang et~al\mbox{.}(2025)]%
        {huang2025m1}
\bibfield{author}{\bibinfo{person}{Xiaoke Huang}, \bibinfo{person}{Juncheng Wu}, \bibinfo{person}{Hui Liu}, \bibinfo{person}{Xianfeng Tang}, {and} \bibinfo{person}{Yuyin Zhou}.} \bibinfo{year}{2025}\natexlab{}.
\newblock \showarticletitle{m1: Unleash the Potential of Test-Time Scaling for Medical Reasoning with Large Language Models}.
\newblock \bibinfo{journal}{\emph{arXiv preprint arXiv:2504.00869}} (\bibinfo{year}{2025}).
\newblock


\bibitem[Jaech et~al\mbox{.}(2024)]%
        {jaech2024openai}
\bibfield{author}{\bibinfo{person}{Aaron Jaech}, \bibinfo{person}{Adam Kalai}, \bibinfo{person}{Adam Lerer}, \bibinfo{person}{Adam Richardson}, \bibinfo{person}{Ahmed El-Kishky}, \bibinfo{person}{Aiden Low}, \bibinfo{person}{Alec Helyar}, \bibinfo{person}{Aleksander Madry}, \bibinfo{person}{Alex Beutel}, \bibinfo{person}{Alex Carney}, {et~al\mbox{.}}} \bibinfo{year}{2024}\natexlab{}.
\newblock \showarticletitle{Openai o1 system card}.
\newblock \bibinfo{journal}{\emph{arXiv preprint arXiv:2412.16720}} (\bibinfo{year}{2024}).
\newblock


\bibitem[Jeon and Lee(2023)]%
        {jeon2023large}
\bibfield{author}{\bibinfo{person}{Jaeho Jeon} {and} \bibinfo{person}{Seongyong Lee}.} \bibinfo{year}{2023}\natexlab{}.
\newblock \showarticletitle{Large language models in education: A focus on the complementary relationship between human teachers and ChatGPT}.
\newblock \bibinfo{journal}{\emph{Education and Information Technologies}} \bibinfo{volume}{28}, \bibinfo{number}{12} (\bibinfo{year}{2023}), \bibinfo{pages}{15873--15892}.
\newblock


\bibitem[Jiang et~al\mbox{.}(2024)]%
        {Slow_Thinking_with_LLMs_1}
\bibfield{author}{\bibinfo{person}{Jinhao Jiang}, \bibinfo{person}{Zhipeng Chen}, \bibinfo{person}{Yingqian Min}, \bibinfo{person}{Jie Chen}, \bibinfo{person}{Xiaoxue Cheng}, \bibinfo{person}{Jiapeng Wang}, \bibinfo{person}{Yiru Tang}, \bibinfo{person}{Haoxiang Sun}, \bibinfo{person}{Jia Deng}, \bibinfo{person}{Wayne~Xin Zhao}, \bibinfo{person}{Zheng Liu}, \bibinfo{person}{Dong Yan}, \bibinfo{person}{Jian Xie}, \bibinfo{person}{Zhongyuan Wang}, {and} \bibinfo{person}{Ji-Rong Wen}.} \bibinfo{year}{2024}\natexlab{}.
\newblock \showarticletitle{Enhancing LLM Reasoning with Reward-guided Tree Search}.
\newblock \bibinfo{journal}{\emph{arXiv preprint arXiv:2411.11694}} (\bibinfo{year}{2024}).
\newblock


\bibitem[Jung et~al\mbox{.}(2024)]%
        {jung2024clst}
\bibfield{author}{\bibinfo{person}{Heeseok Jung}, \bibinfo{person}{Jaesang Yoo}, \bibinfo{person}{Yohaan Yoon}, {and} \bibinfo{person}{Yeonju Jang}.} \bibinfo{year}{2024}\natexlab{}.
\newblock \showarticletitle{CLST: Cold-Start Mitigation in Knowledge Tracing by Aligning a Generative Language Model as a Students' Knowledge Tracer}.
\newblock \bibinfo{journal}{\emph{arXiv preprint arXiv:2406.10296}} (\bibinfo{year}{2024}).
\newblock


\bibitem[Jurenka et~al\mbox{.}(2024a)]%
        {jurenka2024responsibledevelopmentgenerativeai}
\bibfield{author}{\bibinfo{person}{Irina Jurenka}, \bibinfo{person}{Markus Kunesch}, \bibinfo{person}{Kevin~R. McKee}, \bibinfo{person}{Daniel Gillick}, \bibinfo{person}{Shaojian Zhu}, \bibinfo{person}{... Sara~Wiltberger}, {and} \bibinfo{person}{Lila Ibrahim}.} \bibinfo{year}{2024}\natexlab{a}.
\newblock \bibinfo{title}{Towards Responsible Development of Generative AI for Education: An Evaluation-Driven Approach}.
\newblock
\showeprint[arxiv]{2407.12687}~[cs.CY]
\urldef\tempurl%
\url{https://arxiv.org/abs/2407.12687}
\showURL{%
\tempurl}


\bibitem[Jurenka et~al\mbox{.}(2024b)]%
        {jurenka2024learnlm}
\bibfield{author}{\bibinfo{person}{Irina Jurenka}, \bibinfo{person}{Markus Kunesch}, \bibinfo{person}{Kevin~R. McKee}, \bibinfo{person}{Daniel Gillick}, \bibinfo{person}{Shaojian Zhu}, \bibinfo{person}{Sara Wiltberger}, \bibinfo{person}{Shubham~Milind Phal}, \bibinfo{person}{Katherine Hermann}, \bibinfo{person}{Daniel Kasenberg}, \bibinfo{person}{Avishkar Bhoopchand}, \bibinfo{person}{Ankit Anand}, \bibinfo{person}{Miruna Pîslar}, \bibinfo{person}{Stephanie Chan}, \bibinfo{person}{Lisa Wang}, \bibinfo{person}{Jennifer She}, \bibinfo{person}{Parsa Mahmoudieh}, \bibinfo{person}{Aliya Rysbek}, \bibinfo{person}{Wei-Jen Ko}, \bibinfo{person}{Andrea Huber}, \bibinfo{person}{Brett Wiltshire}, \bibinfo{person}{Gal Elidan}, \bibinfo{person}{Roni Rabin}, \bibinfo{person}{Jasmin Rubinovitz}, \bibinfo{person}{Amit Pitaru}, \bibinfo{person}{Mac McAllister}, \bibinfo{person}{Julia Wilkowski}, \bibinfo{person}{David Choi}, \bibinfo{person}{Roee Engelberg}, \bibinfo{person}{Lidan Hackmon}, \bibinfo{person}{Adva Levin},
  \bibinfo{person}{Rachel Griffin}, \bibinfo{person}{Michael Sears}, \bibinfo{person}{Filip Bar}, \bibinfo{person}{Mia Mesar}, \bibinfo{person}{Mana Jabbour}, \bibinfo{person}{Arslan Chaudhry}, \bibinfo{person}{James Cohan}, \bibinfo{person}{Sridhar Thiagarajan}, \bibinfo{person}{Nir Levine}, \bibinfo{person}{Ben Brown}, \bibinfo{person}{Dilan Gorur}, \bibinfo{person}{Svetlana Grant}, \bibinfo{person}{Rachel Hashimshoni}, \bibinfo{person}{Laura Weidinger}, \bibinfo{person}{Jieru Hu}, \bibinfo{person}{Dawn Chen}, \bibinfo{person}{Kuba Dolecki}, \bibinfo{person}{Canfer Akbulut}, \bibinfo{person}{Maxwell Bileschi}, \bibinfo{person}{Laura Culp}, \bibinfo{person}{Wen-Xin Dong}, \bibinfo{person}{Nahema Marchal}, \bibinfo{person}{Kelsie~Van Deman}, \bibinfo{person}{Hema~Bajaj Misra}, \bibinfo{person}{Michael Duah}, \bibinfo{person}{Moran Ambar}, \bibinfo{person}{Avi Caciularu}, \bibinfo{person}{Sandra Lefdal}, \bibinfo{person}{Chris Summerfield}, \bibinfo{person}{James An}, \bibinfo{person}{Pierre-Alexandre
  Kamienny}, \bibinfo{person}{Abhinit Mohdi}, \bibinfo{person}{Theofilos Strinopoulous}, \bibinfo{person}{Annie Hale}, \bibinfo{person}{Wayne Anderson}, \bibinfo{person}{Luis~C. Cobo}, \bibinfo{person}{Niv Efron}, \bibinfo{person}{Muktha Ananda}, \bibinfo{person}{Shakir Mohamed}, \bibinfo{person}{Maureen Heymans}, \bibinfo{person}{Zoubin Ghahramani}, \bibinfo{person}{Yossi Matias}, \bibinfo{person}{Ben Gomes}, {and} \bibinfo{person}{Lila Ibrahim}.} \bibinfo{year}{2024}\natexlab{b}.
\newblock \bibinfo{booktitle}{\emph{Towards Responsible Development of Generative AI for Education: An Evaluation-Driven Approach}}.
\newblock \bibinfo{type}{{T}echnical {R}eport}. \bibinfo{institution}{Google DeepMind}.
\newblock
\urldef\tempurl%
\url{https://goo.gle/LearnLM}
\showURL{%
\tempurl}
\newblock
\shownote{Technical Report}.


\bibitem[Kasneci et~al\mbox{.}(2023)]%
        {kasneci2023chatgpt}
\bibfield{author}{\bibinfo{person}{Enkelejda Kasneci}, \bibinfo{person}{Kathrin Se{\ss}ler}, \bibinfo{person}{Stefan K{\"u}chemann}, \bibinfo{person}{Maria Bannert}, \bibinfo{person}{Daryna Dementieva}, \bibinfo{person}{Frank Fischer}, \bibinfo{person}{Urs Gasser}, \bibinfo{person}{Georg Groh}, \bibinfo{person}{Stephan G{\"u}nnemann}, \bibinfo{person}{Eyke H{\"u}llermeier}, {et~al\mbox{.}}} \bibinfo{year}{2023}\natexlab{}.
\newblock \showarticletitle{ChatGPT for good? On opportunities and challenges of large language models for education}.
\newblock \bibinfo{journal}{\emph{Learning and individual differences}}  \bibinfo{volume}{103} (\bibinfo{year}{2023}), \bibinfo{pages}{102274}.
\newblock


\bibitem[Kim et~al\mbox{.}(2025)]%
        {kim2025limitations}
\bibfield{author}{\bibinfo{person}{Jonathan Kim}, \bibinfo{person}{Anna Podlasek}, \bibinfo{person}{Kie Shidara}, \bibinfo{person}{Feng Liu}, \bibinfo{person}{Ahmed Alaa}, {and} \bibinfo{person}{Danilo Bernardo}.} \bibinfo{year}{2025}\natexlab{}.
\newblock \showarticletitle{Limitations of Large Language Models in Clinical Problem-Solving Arising from Inflexible Reasoning}.
\newblock \bibinfo{journal}{\emph{arXiv preprint arXiv:2502.04381}} (\bibinfo{year}{2025}).
\newblock


\bibitem[Lai et~al\mbox{.}(2025)]%
        {lai2025med}
\bibfield{author}{\bibinfo{person}{Yuxiang Lai}, \bibinfo{person}{Jike Zhong}, \bibinfo{person}{Ming Li}, \bibinfo{person}{Shitian Zhao}, {and} \bibinfo{person}{Xiaofeng Yang}.} \bibinfo{year}{2025}\natexlab{}.
\newblock \showarticletitle{Med-r1: Reinforcement learning for generalizable medical reasoning in vision-language models}.
\newblock \bibinfo{journal}{\emph{arXiv preprint arXiv:2503.13939}} (\bibinfo{year}{2025}).
\newblock


\bibitem[Lee et~al\mbox{.}(2024a)]%
        {lee2024language}
\bibfield{author}{\bibinfo{person}{Unggi Lee}, \bibinfo{person}{Jiyeong Bae}, \bibinfo{person}{Dohee Kim}, \bibinfo{person}{Sookbun Lee}, \bibinfo{person}{Jaekwon Park}, \bibinfo{person}{Taekyung Ahn}, \bibinfo{person}{Gunho Lee}, \bibinfo{person}{Damji Stratton}, {and} \bibinfo{person}{Hyeoncheol Kim}.} \bibinfo{year}{2024}\natexlab{a}.
\newblock \showarticletitle{Language model can do knowledge tracing: Simple but effective method to integrate language model and knowledge tracing task}.
\newblock \bibinfo{journal}{\emph{arXiv preprint arXiv:2406.02893}} (\bibinfo{year}{2024}).
\newblock


\bibitem[Lee et~al\mbox{.}(2024b)]%
        {Lee_2024}
\bibfield{author}{\bibinfo{person}{Unggi Lee}, \bibinfo{person}{Minji Jeon}, \bibinfo{person}{Yunseo Lee}, \bibinfo{person}{Gyuri Byun}, \bibinfo{person}{Yoorim Son}, \bibinfo{person}{Jaeyoon Shin}, \bibinfo{person}{Hongkyu Ko}, {and} \bibinfo{person}{Hyeoncheol Kim}.} \bibinfo{year}{2024}\natexlab{b}.
\newblock \showarticletitle{LLaVA-docent: Instruction tuning with multimodal large language model to support art appreciation education}.
\newblock \bibinfo{journal}{\emph{Computers and Education: Artificial Intelligence}}  \bibinfo{volume}{7} (\bibinfo{date}{Dec.} \bibinfo{year}{2024}), \bibinfo{pages}{100297}.
\newblock
\showISSN{2666-920X}
\href{https://doi.org/10.1016/j.caeai.2024.100297}{doi:\nolinkurl{10.1016/j.caeai.2024.100297}}


\bibitem[Lee et~al\mbox{.}(2024c)]%
        {lee2024difficulty}
\bibfield{author}{\bibinfo{person}{Unggi Lee}, \bibinfo{person}{Sungjun Yoon}, \bibinfo{person}{Joon~Seo Yun}, \bibinfo{person}{Kyoungsoo Park}, \bibinfo{person}{Young~Hoon Jung}, \bibinfo{person}{Damji Stratton}, {and} \bibinfo{person}{Hyeoncheol Kim}.} \bibinfo{year}{2024}\natexlab{c}.
\newblock \showarticletitle{Difficulty-Focused Contrastive Learning for Knowledge Tracing with a Large Language Model-Based Difficulty Prediction}. In \bibinfo{booktitle}{\emph{Joint 30th International Conference on Computational Linguistics and 14th International Conference on Language Resources and Evaluation, LREC-COLING 2024}}. European Language Resources Association (ELRA), \bibinfo{pages}{4891--4900}.
\newblock


\bibitem[Li et~al\mbox{.}(2025)]%
        {li2025system}
\bibfield{author}{\bibinfo{person}{Zhong-Zhi Li}, \bibinfo{person}{Duzhen Zhang}, \bibinfo{person}{Ming-Liang Zhang}, \bibinfo{person}{Jiaxin Zhang}, \bibinfo{person}{Zengyan Liu}, \bibinfo{person}{Yuxuan Yao}, \bibinfo{person}{Haotian Xu}, \bibinfo{person}{Junhao Zheng}, \bibinfo{person}{Pei-Jie Wang}, \bibinfo{person}{Xiuyi Chen}, {et~al\mbox{.}}} \bibinfo{year}{2025}\natexlab{}.
\newblock \showarticletitle{From system 1 to system 2: A survey of reasoning large language models}.
\newblock \bibinfo{journal}{\emph{arXiv preprint arXiv:2502.17419}} (\bibinfo{year}{2025}).
\newblock


\bibitem[Liu et~al\mbox{.}(2025)]%
        {liu2025fin}
\bibfield{author}{\bibinfo{person}{Zhaowei Liu}, \bibinfo{person}{Xin Guo}, \bibinfo{person}{Fangqi Lou}, \bibinfo{person}{Lingfeng Zeng}, \bibinfo{person}{Jinyi Niu}, \bibinfo{person}{Zixuan Wang}, \bibinfo{person}{Jiajie Xu}, \bibinfo{person}{Weige Cai}, \bibinfo{person}{Ziwei Yang}, \bibinfo{person}{Xueqian Zhao}, {et~al\mbox{.}}} \bibinfo{year}{2025}\natexlab{}.
\newblock \showarticletitle{Fin-r1: A large language model for financial reasoning through reinforcement learning}.
\newblock \bibinfo{journal}{\emph{arXiv preprint arXiv:2503.16252}} (\bibinfo{year}{2025}).
\newblock


\bibitem[Liu et~al\mbox{.}(2024)]%
        {liu2024xes3g5m}
\bibfield{author}{\bibinfo{person}{Zitao Liu}, \bibinfo{person}{Qiongqiong Liu}, \bibinfo{person}{Teng Guo}, \bibinfo{person}{Jiahao Chen}, \bibinfo{person}{Shuyan Huang}, \bibinfo{person}{Xiangyu Zhao}, \bibinfo{person}{Jiliang Tang}, \bibinfo{person}{Weiqi Luo}, {and} \bibinfo{person}{Jian Weng}.} \bibinfo{year}{2024}\natexlab{}.
\newblock \showarticletitle{XES3G5M: A Knowledge Tracing Benchmark Dataset with Auxiliary Information}.
\newblock \bibinfo{journal}{\emph{Advances in Neural Information Processing Systems}}  \bibinfo{volume}{36} (\bibinfo{year}{2024}).
\newblock


\bibitem[Loughran(2002)]%
        {loughran2002developing}
\bibfield{author}{\bibinfo{person}{J~John Loughran}.} \bibinfo{year}{2002}\natexlab{}.
\newblock \bibinfo{booktitle}{\emph{Developing reflective practice: Learning about teaching and learning through modelling}}.
\newblock \bibinfo{publisher}{Routledge}.
\newblock


\bibitem[Macina et~al\mbox{.}(2025)]%
        {macina2025mathtutorbench}
\bibfield{author}{\bibinfo{person}{Jakub Macina}, \bibinfo{person}{Nico Daheim}, \bibinfo{person}{Ido Hakimi}, \bibinfo{person}{Manu Kapur}, \bibinfo{person}{Iryna Gurevych}, {and} \bibinfo{person}{Mrinmaya Sachan}.} \bibinfo{year}{2025}\natexlab{}.
\newblock \showarticletitle{MathTutorBench: A Benchmark for Measuring Open-ended Pedagogical Capabilities of LLM Tutors}.
\newblock \bibinfo{journal}{\emph{arXiv preprint arXiv:2502.18940}} (\bibinfo{year}{2025}).
\newblock


\bibitem[McGarr(2021)]%
        {mcgarr2021use}
\bibfield{author}{\bibinfo{person}{Oliver McGarr}.} \bibinfo{year}{2021}\natexlab{}.
\newblock \showarticletitle{The use of virtual simulations in teacher education to develop pre-service teachers’ behaviour and classroom management skills: implications for reflective practice}.
\newblock \bibinfo{journal}{\emph{Journal of Education for Teaching}} \bibinfo{volume}{47}, \bibinfo{number}{2} (\bibinfo{year}{2021}), \bibinfo{pages}{274--286}.
\newblock


\bibitem[Min et~al\mbox{.}(2024)]%
        {Slow_Thinking_with_LLMs_2}
\bibfield{author}{\bibinfo{person}{Yingqian Min}, \bibinfo{person}{Zhipeng Chen}, \bibinfo{person}{Jinhao Jiang}, \bibinfo{person}{Jie Chen}, \bibinfo{person}{Jia Deng}, \bibinfo{person}{Yiwen Hu}, \bibinfo{person}{Yiru Tang}, \bibinfo{person}{Jiapeng Wang}, \bibinfo{person}{Xiaoxue Cheng}, \bibinfo{person}{Huatong Song}, \bibinfo{person}{Wayne~Xin Zhao}, \bibinfo{person}{Zheng Liu}, \bibinfo{person}{Zhongyuan Wang}, {and} \bibinfo{person}{Ji-Rong Wen}.} \bibinfo{year}{2024}\natexlab{}.
\newblock \showarticletitle{Imitate, Explore, and Self-Improve: A Reproduction Report on Slow-thinking Reasoning Systems}.
\newblock \bibinfo{journal}{\emph{arXiv preprint arXiv:2412.09413}} (\bibinfo{year}{2024}).
\newblock


\bibitem[Mizumoto and Eguchi(2023)]%
        {mizumoto2023exploring}
\bibfield{author}{\bibinfo{person}{Atsushi Mizumoto} {and} \bibinfo{person}{Masaki Eguchi}.} \bibinfo{year}{2023}\natexlab{}.
\newblock \showarticletitle{Exploring the potential of using an AI language model for automated essay scoring}.
\newblock \bibinfo{journal}{\emph{Research Methods in Applied Linguistics}} \bibinfo{volume}{2}, \bibinfo{number}{2} (\bibinfo{year}{2023}), \bibinfo{pages}{100050}.
\newblock


\bibitem[Nelson(2020)]%
        {nelson2020computational}
\bibfield{author}{\bibinfo{person}{Laura~K Nelson}.} \bibinfo{year}{2020}\natexlab{}.
\newblock \showarticletitle{Computational grounded theory: A methodological framework}.
\newblock \bibinfo{journal}{\emph{Sociological Methods \& Research}} \bibinfo{volume}{49}, \bibinfo{number}{1} (\bibinfo{year}{2020}), \bibinfo{pages}{3--42}.
\newblock


\bibitem[OpenAI(2025)]%
        {openai2025gpt41}
\bibfield{author}{\bibinfo{person}{OpenAI}.} \bibinfo{year}{2025}\natexlab{}.
\newblock \bibinfo{booktitle}{\emph{Introducing GPT-4.1 in the API}}.
\newblock
\urldef\tempurl%
\url{https://openai.com/index/gpt-4-1/}
\showURL{%
\tempurl}
\newblock
\shownote{Accessed: 2025-05-17}.


\bibitem[Pan et~al\mbox{.}(2025)]%
        {pan2025medvlm}
\bibfield{author}{\bibinfo{person}{Jiazhen Pan}, \bibinfo{person}{Che Liu}, \bibinfo{person}{Junde Wu}, \bibinfo{person}{Fenglin Liu}, \bibinfo{person}{Jiayuan Zhu}, \bibinfo{person}{Hongwei~Bran Li}, \bibinfo{person}{Chen Chen}, \bibinfo{person}{Cheng Ouyang}, {and} \bibinfo{person}{Daniel Rueckert}.} \bibinfo{year}{2025}\natexlab{}.
\newblock \showarticletitle{Medvlm-r1: Incentivizing medical reasoning capability of vision-language models (vlms) via reinforcement learning}.
\newblock \bibinfo{journal}{\emph{arXiv preprint arXiv:2502.19634}} (\bibinfo{year}{2025}).
\newblock


\bibitem[Puech et~al\mbox{.}(2025)]%
        {puech2025pedagogicalsteering}
\bibfield{author}{\bibinfo{person}{Romain Puech}, \bibinfo{person}{Jakub Macina}, \bibinfo{person}{Julia Chatain}, \bibinfo{person}{Mrinmaya Sachan}, {and} \bibinfo{person}{Manu Kapur}.} \bibinfo{year}{2025}\natexlab{}.
\newblock \showarticletitle{Towards the Pedagogical Steering of Large Language Models for Tutoring: A Case Study with Modeling Productive Failure}.
\newblock \bibinfo{journal}{\emph{arXiv preprint arXiv:2410.03781}} (\bibinfo{year}{2025}).
\newblock


\bibitem[Rasul et~al\mbox{.}(2023)]%
        {rasul2023role}
\bibfield{author}{\bibinfo{person}{Tareq Rasul}, \bibinfo{person}{Sumesh Nair}, \bibinfo{person}{Diane Kalendra}, \bibinfo{person}{Mulyadi Robin}, \bibinfo{person}{Fernando de Oliveira~Santini}, \bibinfo{person}{Wagner~Junior Ladeira}, \bibinfo{person}{Mingwei Sun}, \bibinfo{person}{Ingrid Day}, \bibinfo{person}{Raouf~Ahmad Rather}, {and} \bibinfo{person}{Liz Heathcote}.} \bibinfo{year}{2023}\natexlab{}.
\newblock \showarticletitle{The role of ChatGPT in higher education: Benefits, challenges, and future research directions}.
\newblock \bibinfo{journal}{\emph{Journal of Applied Learning and Teaching}} \bibinfo{volume}{6}, \bibinfo{number}{1} (\bibinfo{year}{2023}), \bibinfo{pages}{41--56}.
\newblock


\bibitem[Roberts et~al\mbox{.}(2024)]%
        {roberts2024artificial}
\bibfield{author}{\bibinfo{person}{John Roberts}, \bibinfo{person}{Max Baker}, {and} \bibinfo{person}{Jane Andrew}.} \bibinfo{year}{2024}\natexlab{}.
\newblock \showarticletitle{Artificial intelligence and qualitative research: The promise and perils of large language model (LLM)‘assistance’}.
\newblock \bibinfo{journal}{\emph{Critical Perspectives on Accounting}}  \bibinfo{volume}{99} (\bibinfo{year}{2024}), \bibinfo{pages}{102722}.
\newblock


\bibitem[Sch{\"o}n(2017)]%
        {schon2017reflective}
\bibfield{author}{\bibinfo{person}{Donald~A Sch{\"o}n}.} \bibinfo{year}{2017}\natexlab{}.
\newblock \showarticletitle{The reflective practitioner: How professionals think in action}.
\newblock  (\bibinfo{year}{2017}).
\newblock


\bibitem[Siuty et~al\mbox{.}(2018)]%
        {siuty2018unraveling}
\bibfield{author}{\bibinfo{person}{Molly~Baustien Siuty}, \bibinfo{person}{Melinda~M Leko}, {and} \bibinfo{person}{Kimberly~M Knackstedt}.} \bibinfo{year}{2018}\natexlab{}.
\newblock \showarticletitle{Unraveling the role of curriculum in teacher decision making}.
\newblock \bibinfo{journal}{\emph{Teacher Education and Special Education}} \bibinfo{volume}{41}, \bibinfo{number}{1} (\bibinfo{year}{2018}), \bibinfo{pages}{39--57}.
\newblock


\bibitem[Slade et~al\mbox{.}(2019)]%
        {slade2019impact}
\bibfield{author}{\bibinfo{person}{Mary~L Slade}, \bibinfo{person}{Tammy~J Burnham}, \bibinfo{person}{Sarah~Marie Catalana}, {and} \bibinfo{person}{Tammy Waters}.} \bibinfo{year}{2019}\natexlab{}.
\newblock \showarticletitle{The Impact of Reflective Practice on Teacher Candidates' Learning.}
\newblock \bibinfo{journal}{\emph{International Journal for the Scholarship of Teaching and Learning}} \bibinfo{volume}{13}, \bibinfo{number}{2} (\bibinfo{year}{2019}), \bibinfo{pages}{15}.
\newblock


\bibitem[Sonkar et~al\mbox{.}(2024)]%
        {sonkar2024pedagogicalalignment}
\bibfield{author}{\bibinfo{person}{Shashank Sonkar}, \bibinfo{person}{Kangqi Ni}, \bibinfo{person}{Sapana Chaudhary}, {and} \bibinfo{person}{Richard~G. Baraniuk}.} \bibinfo{year}{2024}\natexlab{}.
\newblock \showarticletitle{Pedagogical Alignment of Large Language Models}.
\newblock \bibinfo{journal}{\emph{arXiv preprint arXiv:2402.05000}} (\bibinfo{year}{2024}).
\newblock


\bibitem[Team(2025a)]%
        {sky_t1_2025}
\bibfield{author}{\bibinfo{person}{NovaSky Team}.} \bibinfo{year}{2025}\natexlab{a}.
\newblock \bibinfo{title}{Sky-T1: Train your own O1 preview model within \$450}.
\newblock \bibinfo{howpublished}{https://novasky-ai.github.io/posts/sky-t1}.
\newblock
\newblock
\shownote{Accessed: 2025-01-09}.


\bibitem[Team(2025b)]%
        {qwq32b}
\bibfield{author}{\bibinfo{person}{Qwen Team}.} \bibinfo{year}{2025}\natexlab{b}.
\newblock \bibinfo{title}{QwQ-32B: Embracing the Power of Reinforcement Learning}.
\newblock
\urldef\tempurl%
\url{https://qwenlm.github.io/blog/qwq-32b/}
\showURL{%
\tempurl}


\bibitem[Treacy and Gaunt(2021)]%
        {treacy2021promoting}
\bibfield{author}{\bibinfo{person}{Danielle Treacy} {and} \bibinfo{person}{Helena Gaunt}.} \bibinfo{year}{2021}\natexlab{}.
\newblock \showarticletitle{Promoting interconnections between reflective practice and collective creativity in higher arts education: the potential of engaging with a reflective matrix}.
\newblock \bibinfo{journal}{\emph{Reflective Practice}} \bibinfo{volume}{22}, \bibinfo{number}{4} (\bibinfo{year}{2021}), \bibinfo{pages}{488--500}.
\newblock


\bibitem[Wang et~al\mbox{.}(2024)]%
        {wang2024large}
\bibfield{author}{\bibinfo{person}{Shen Wang}, \bibinfo{person}{Tianlong Xu}, \bibinfo{person}{Hang Li}, \bibinfo{person}{Chaoli Zhang}, \bibinfo{person}{Joleen Liang}, \bibinfo{person}{Jiliang Tang}, \bibinfo{person}{Philip~S Yu}, {and} \bibinfo{person}{Qingsong Wen}.} \bibinfo{year}{2024}\natexlab{}.
\newblock \showarticletitle{Large language models for education: A survey and outlook}.
\newblock \bibinfo{journal}{\emph{arXiv preprint arXiv:2403.18105}} (\bibinfo{year}{2024}).
\newblock


\bibitem[Wang et~al\mbox{.}(2025)]%
        {wang2025thoughts}
\bibfield{author}{\bibinfo{person}{Yue Wang}, \bibinfo{person}{Qiuzhi Liu}, \bibinfo{person}{Jiahao Xu}, \bibinfo{person}{Tian Liang}, \bibinfo{person}{Xingyu Chen}, \bibinfo{person}{Zhiwei He}, \bibinfo{person}{Linfeng Song}, \bibinfo{person}{Dian Yu}, \bibinfo{person}{Juntao Li}, \bibinfo{person}{Zhuosheng Zhang}, {et~al\mbox{.}}} \bibinfo{year}{2025}\natexlab{}.
\newblock \showarticletitle{Thoughts Are All Over the Place: On the Underthinking of o1-Like LLMs}.
\newblock \bibinfo{journal}{\emph{arXiv preprint arXiv:2501.18585}} (\bibinfo{year}{2025}).
\newblock


\bibitem[Whalen et~al\mbox{.}(2023)]%
        {whalen2023chatgpt}
\bibfield{author}{\bibinfo{person}{Jeromie Whalen}, \bibinfo{person}{Chrystalla Mouza}, {et~al\mbox{.}}} \bibinfo{year}{2023}\natexlab{}.
\newblock \showarticletitle{ChatGPT: Challenges, opportunities, and implications for teacher education}.
\newblock \bibinfo{journal}{\emph{Contemporary Issues in Technology and Teacher Education}} \bibinfo{volume}{23}, \bibinfo{number}{1} (\bibinfo{year}{2023}), \bibinfo{pages}{1--23}.
\newblock


\bibitem[Xu et~al\mbox{.}(2025)]%
        {xu2025towards}
\bibfield{author}{\bibinfo{person}{Fengli Xu}, \bibinfo{person}{Qianyue Hao}, \bibinfo{person}{Zefang Zong}, \bibinfo{person}{Jingwei Wang}, \bibinfo{person}{Yunke Zhang}, \bibinfo{person}{Jingyi Wang}, \bibinfo{person}{Xiaochong Lan}, \bibinfo{person}{Jiahui Gong}, \bibinfo{person}{Tianjian Ouyang}, \bibinfo{person}{Fanjin Meng}, {et~al\mbox{.}}} \bibinfo{year}{2025}\natexlab{}.
\newblock \showarticletitle{Towards Large Reasoning Models: A Survey of Reinforced Reasoning with Large Language Models}.
\newblock \bibinfo{journal}{\emph{arXiv preprint arXiv:2501.09686}} (\bibinfo{year}{2025}).
\newblock


\bibitem[Yan et~al\mbox{.}(2024)]%
        {yan2024practical}
\bibfield{author}{\bibinfo{person}{Lixiang Yan}, \bibinfo{person}{Lele Sha}, \bibinfo{person}{Linxuan Zhao}, \bibinfo{person}{Yuheng Li}, \bibinfo{person}{Roberto Martinez-Maldonado}, \bibinfo{person}{Guanliang Chen}, \bibinfo{person}{Xinyu Li}, \bibinfo{person}{Yueqiao Jin}, {and} \bibinfo{person}{Dragan Ga{\v{s}}evi{\'c}}.} \bibinfo{year}{2024}\natexlab{}.
\newblock \showarticletitle{Practical and ethical challenges of large language models in education: A systematic scoping review}.
\newblock \bibinfo{journal}{\emph{British Journal of Educational Technology}} \bibinfo{volume}{55}, \bibinfo{number}{1} (\bibinfo{year}{2024}), \bibinfo{pages}{90--112}.
\newblock


\bibitem[Yang et~al\mbox{.}(2024)]%
        {qwen2.5}
\bibfield{author}{\bibinfo{person}{An Yang}, \bibinfo{person}{Baosong Yang}, \bibinfo{person}{Beichen Zhang}, \bibinfo{person}{Binyuan Hui}, \bibinfo{person}{Bo Zheng}, \bibinfo{person}{Bowen Yu}, \bibinfo{person}{Chengyuan Li}, \bibinfo{person}{Dayiheng Liu}, \bibinfo{person}{Fei Huang}, \bibinfo{person}{Haoran Wei}, \bibinfo{person}{Huan Lin}, \bibinfo{person}{Jian Yang}, \bibinfo{person}{Jianhong Tu}, \bibinfo{person}{Jianwei Zhang}, \bibinfo{person}{Jianxin Yang}, \bibinfo{person}{Jiaxi Yang}, \bibinfo{person}{Jingren Zhou}, \bibinfo{person}{Junyang Lin}, \bibinfo{person}{Kai Dang}, \bibinfo{person}{Keming Lu}, \bibinfo{person}{Keqin Bao}, \bibinfo{person}{Kexin Yang}, \bibinfo{person}{Le Yu}, \bibinfo{person}{Mei Li}, \bibinfo{person}{Mingfeng Xue}, \bibinfo{person}{Pei Zhang}, \bibinfo{person}{Qin Zhu}, \bibinfo{person}{Rui Men}, \bibinfo{person}{Runji Lin}, \bibinfo{person}{Tianhao Li}, \bibinfo{person}{Tianyi Tang}, \bibinfo{person}{Tingyu Xia}, \bibinfo{person}{Xingzhang Ren},
  \bibinfo{person}{Xuancheng Ren}, \bibinfo{person}{Yang Fan}, \bibinfo{person}{Yang Su}, \bibinfo{person}{Yichang Zhang}, \bibinfo{person}{Yu Wan}, \bibinfo{person}{Yuqiong Liu}, \bibinfo{person}{Zeyu Cui}, \bibinfo{person}{Zhenru Zhang}, {and} \bibinfo{person}{Zihan Qiu}.} \bibinfo{year}{2024}\natexlab{}.
\newblock \showarticletitle{Qwen2.5 Technical Report}.
\newblock \bibinfo{journal}{\emph{arXiv preprint arXiv:2412.15115}} (\bibinfo{year}{2024}).
\newblock


\bibitem[Yu et~al\mbox{.}(2025)]%
        {yu2025evaluating}
\bibfield{author}{\bibinfo{person}{Yaoyao Yu}, \bibinfo{person}{Leilei Gan}, \bibinfo{person}{Yinghao Hu}, \bibinfo{person}{Bin Wei}, \bibinfo{person}{Kun Kuang}, {and} \bibinfo{person}{Fei Wu}.} \bibinfo{year}{2025}\natexlab{}.
\newblock \showarticletitle{Evaluating test-time scaling llms for legal reasoning: Openai o1, deepseek-r1, and beyond}.
\newblock \bibinfo{journal}{\emph{arXiv preprint arXiv:2503.16040}} (\bibinfo{year}{2025}).
\newblock


\bibitem[Zhan et~al\mbox{.}(2024)]%
        {zhan2024knowledge}
\bibfield{author}{\bibinfo{person}{Bojun Zhan}, \bibinfo{person}{Teng Guo}, \bibinfo{person}{Xueyi Li}, \bibinfo{person}{Mingliang Hou}, \bibinfo{person}{Qianru Liang}, \bibinfo{person}{Boyu Gao}, \bibinfo{person}{Weiqi Luo}, {and} \bibinfo{person}{Zitao Liu}.} \bibinfo{year}{2024}\natexlab{}.
\newblock \showarticletitle{Knowledge tracing as language processing: a large-scale autoregressive paradigm}. In \bibinfo{booktitle}{\emph{International Conference on Artificial Intelligence in Education}}. Springer, \bibinfo{pages}{177--191}.
\newblock


\bibitem[Zhang et~al\mbox{.}(2024)]%
        {zhang2024cjeval}
\bibfield{author}{\bibinfo{person}{Qian-Wen Zhang}, \bibinfo{person}{Haochen Wang}, \bibinfo{person}{Fang Li}, \bibinfo{person}{Siyu An}, \bibinfo{person}{Lingfeng Qiao}, \bibinfo{person}{Liangcai Gao}, \bibinfo{person}{Di Yin}, {and} \bibinfo{person}{Xing Sun}.} \bibinfo{year}{2024}\natexlab{}.
\newblock \showarticletitle{CJEval: A Benchmark for Assessing Large Language Models Using Chinese Junior High School Exam Data}.
\newblock  (\bibinfo{year}{2024}).
\newblock
\showeprint[arxiv]{2409.16202}~[cs.AI]


\bibitem[Zhang et~al\mbox{.}(2025)]%
        {zhang2025med}
\bibfield{author}{\bibinfo{person}{Sheng Zhang}, \bibinfo{person}{Qianchu Liu}, \bibinfo{person}{Guanghui Qin}, \bibinfo{person}{Tristan Naumann}, {and} \bibinfo{person}{Hoifung Poon}.} \bibinfo{year}{2025}\natexlab{}.
\newblock \showarticletitle{Med-RLVR: Emerging Medical Reasoning from a 3B base model via reinforcement Learning}.
\newblock \bibinfo{journal}{\emph{arXiv preprint arXiv:2502.19655}} (\bibinfo{year}{2025}).
\newblock


\end{thebibliography}
\end{document}